\documentclass[journal,journal,comsoc]{IEEEtran}

\ifCLASSOPTIONcompsoc
  \usepackage[nocompress]{cite}
\else
  \usepackage{cite}
\fi
\ifCLASSINFOpdf
\else
\fi

\usepackage{times}
\usepackage{epsfig}
\usepackage{graphicx}
\usepackage{amsmath}
\usepackage{amssymb}
\usepackage{footnote}
\usepackage{dsfont}
\usepackage{enumitem}

\usepackage{comment}
\usepackage{blindtext}
\usepackage{morewrites}
\usepackage{makecell}
\usepackage{pifont}
\newcommand{\xmark}{\ding{55}}%
\newcommand{\cmark}{\ding{51}}%
\usepackage{multirow}
\usepackage{xcolor}
\usepackage{float}
\usepackage{dblfloatfix}

\begin{document}

\title{CoMaL: Conditional Maximum Likelihood Approach to Self-supervised Domain Adaptation in Long-tail Semantic Segmentation}

\author{
Thanh-Dat Truong$^1$~\IEEEmembership{Student~Member,~IEEE},
Chi Nhan Duong$^2$~\IEEEmembership{Member,~IEEE},
Pierce Helton$^1$, %
Ashley~Dowling$^3$, 
Xin Li$^4$~\IEEEmembership{Fellow,~IEEE},
Khoa Luu$^1$~\IEEEmembership{Member,~IEEE},
\thanks{$^{1}$Department of Computer Science and Computer Engineering , University of Arkansas;
$^{2}$Department of Computer Science and Software Engineering, Concordia University, Concordia University;
$^{3}$Department of Entomology and Plant Pathology, University of Arkansas; \emph{and} 
$^{4}$Lane Department of Computer Science and Electrical Engineering, West Virginia University
Emails: \textit{tt032@uark.edu, dcnhan@ieee.org, pchelton@email.uark.edu, adowling@uark.edu, Xin.Li@mail.wvu.edu, khoaluu@uark.edu}}
}

\IEEEtitleabstractindextext{%
\begin{abstract}
The research in self-supervised domain adaptation in semantic segmentation has recently received considerable attention. 
Although GAN-based methods have become one of the most popular approaches to domain adaptation, they have suffered from some limitations. They are insufficient to model both global and local structures of a given image, especially in small regions of tail classes. Moreover, they perform bad on the tail classes containing limited number of pixels or less training samples.
In order to address these issues, we present a new self-supervised domain adaptation approach to tackle long-tail semantic segmentation in this paper. Firstly, a new metric is introduced to formulate long-tail domain adaptation in the segmentation problem. 
Secondly, a new Conditional Maximum Likelihood (CoMaL) approach\footnote{The source code implementation of CoMaL will be publicly available.} in an autoregressive framework is presented to solve the problem of long-tail domain adaptation. Although other segmentation methods work under the pixel independence assumption, the long-tailed pixel distributions in CoMaL are generally solved in the context of structural dependency, as that is more realistic.
Finally, the proposed method is evaluated on popular large-scale semantic segmentation benchmarks, i.e., \textit{``SYNTHIA $\to$ Cityscapes''} and \textit{``GTA $\to$ Cityscapes''},
and outperforms the prior methods by a large margin in both the standard and the proposed evaluation protocols.
\end{abstract}

}

\maketitle

\IEEEdisplaynontitleabstractindextext

\IEEEpeerreviewmaketitle

\vspace{10mm}
\IEEEraisesectionheading{\section{Introduction}\label{sec:intro}}

Semantic scene segmentation has become one of the most popular topics in computer vision. It aims to densely assign each pixel in a given image to the corresponding predefined class. 
Recently, deep learning-based approaches have achieved remarkable results in semantic segmentation \cite{chen2018deeplab, chen2017rethinking, xie2021segformer}. 
A typical segmentation method is usually trained on scene datasets with labels. However, annotating images for the semantic segmentation task is costly and time-consuming, since it requires every single pixel in a given image to be labeled. Another approach of reducing the cost of annotating images is to use a simulation to create a large-scale synthetic dataset \cite{Richter_2016_ECCV, Ros_2016_CVPR}. However, deploying the supervised models trained on synthetic datasets to real images is not an appropriate solution, since these supervised models often perform worse on the real images because of a pixel appearance gap between synthetic and real images.

\begin{figure}[!t]
    \centering
    \includegraphics[width=0.5\textwidth]{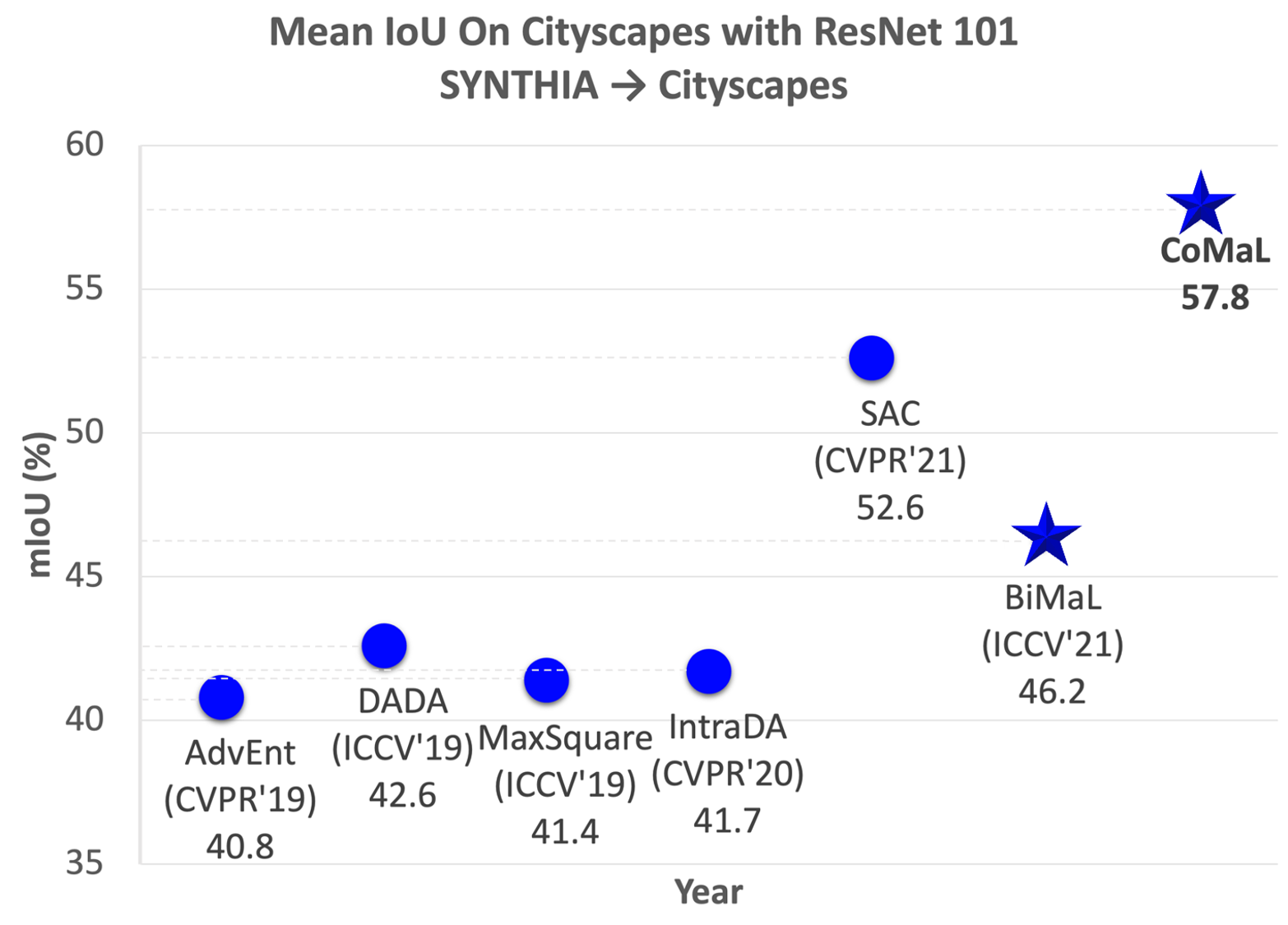}
    \caption{\textbf{Overview of Results.} The figure illustrates the mIoU results on Cityscapes. Our CoMaL and BiMaL approach achieves the State-of-the-Art results and outperforms prior methods by a large margin.} 
    \label{fig:first_fig}
\end{figure}

\begin{figure*}[!t]
    \centering
    \includegraphics[width=1.0\textwidth]{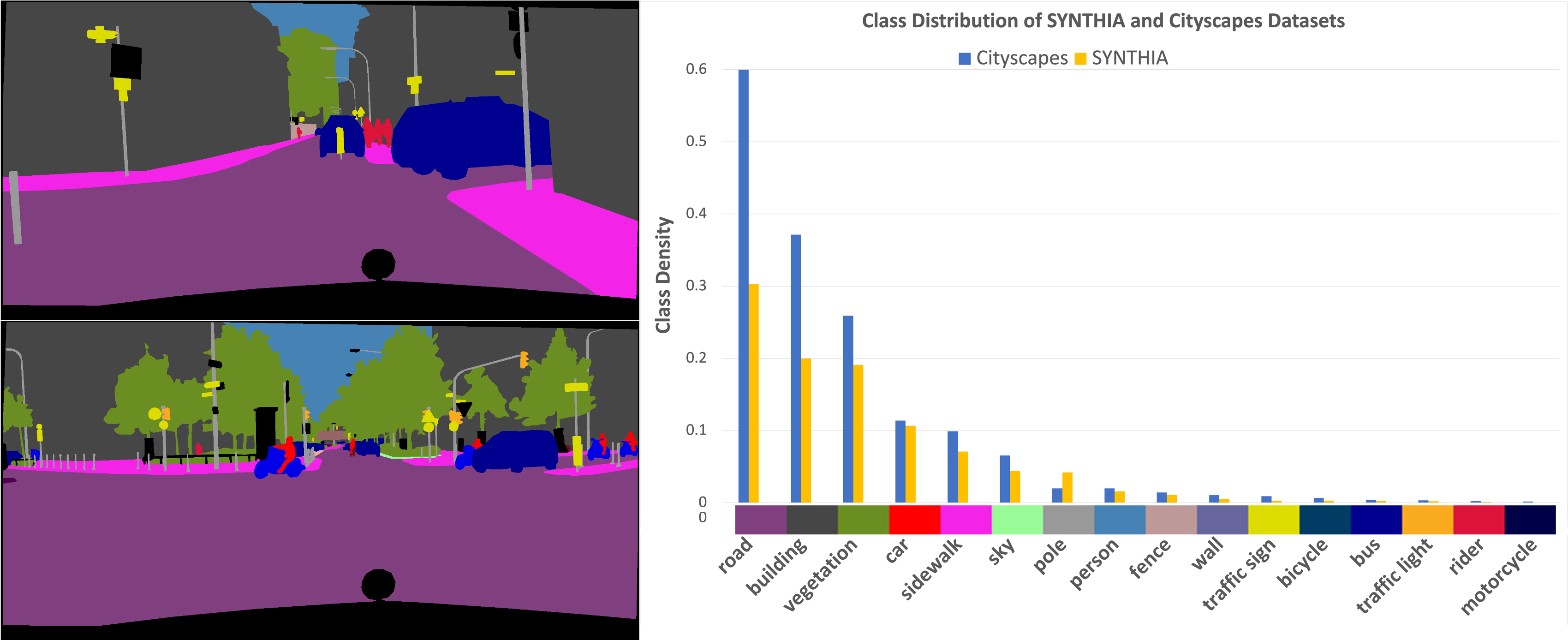}
    \caption{\textbf{The class distributions of SYNTHIA and Cityscapes datasets.} The left figure illustrates the typical segmentation ground-truths of the Cityscapes. The right image show the class distribution over the entire datasets of Cityscapes and SYNTHIA, respectively. Best view in color.}
    \label{fig:long_tailed_distribition}
\end{figure*}
Self-supervised Domain Adaptation (SDA) aims to learn a model on the annotated source datasets and adapt to unlabeled target datasets to guarantee its performance on the new domain. 
Common domain adaptation methods usually minimize the distribution discrepancy of the deep representations extracted from the source and target domains in addition to supervised training on the source domain \cite{tsai2019domain, tsai2018learning, vu2019advent, vu2019dada}. The discrepancy minimization can be processed at single or multiple levels of deep feature representation using 
maximum mean discrepancy \cite{ganin2015unsupervised, long2015learning}
, adversarial training \cite{chen2018road, hoffman18a, hong2018CVPR, tsai2018learning, tzeng2017adversarial}, %
or contrastive learning \cite{Yue_2021_CVPR, kang2019contrastive}.
They have shown their potential performance in aligning the predicted outputs across domains. 
Another self-training approach that %
utilizes pseudo labels \cite{zhang2021prototypical, Araslanov:2021:DASAC, zou2018unsupervised} on the target domain has also drawn much attention in recent years. 

Although these prior methods have shown their high performance in domain adaptation for semantic segmentation, they are still unable to perform well on the tail classes that have limited numbers of pixels, i.e. intra long-tail classes, or less training samples, i.e. inter long-tail classes. 
As shown in Fig. \ref{fig:long_tailed_distribition}, the number of pixels in classes such as bicycles, traffic signs, and traffic light are limited compared to other classes like road, building, and sky. Although the performance of the trained model on the tailed class in the source domain is reasonable, its performance on the tailed classes in the target domain dramatically drops 
(see Table \ref{tab:synthia2city}).
In adversarial methods \cite{hoffman2016fcns, tsai2018learning, tsai2019domain}, the learned discriminator usually provides a weak indication of structural learning for the semantic segmentation due to the binary cross-entropy predictions between source and target domains.
As a result, they are insufficient in the global and the local structures of a given image, especially in small regions of the tail class. Another entropy metric has been proposed in self-training methods \cite{vu2019advent, pan2020unsupervised} to improve the prediction confidence. 
However, this metric tends to highly response to the head classes that are usually occupy a lot of pixels in an image rather than tail classes which contribute very limited number of pixels. 
More importantly, an assumption of \textit{pixel independence} is required for the metric to be applicable.

\noindent
\textbf{Contributions of This Work.}
This work presents a new self-supervised domain adaptation approach to long-tail semantic scene segmentation,  which is an extension of our previous work \cite{truong2021bimal}. In that work, we introduced a new Unaligned Domain Score to measure the efficiency of a learned
model on a new target domain in unsupervised manner, and presented
a new Bijective Maximum Likelihood
(BiMaL) loss that is a generalized form of the Adversarial
Entropy Minimization without any assumption about pixel
independence.
In this work, we further address the long-tail issue that occurs in both labeled source domain \textbf{and} new \textbf{\textit{unlabeled}} target domains (as shown in Fig. \ref{fig:long_tailed_distribition}). 
Unlike prior works where only labeled part is on their focus to alleviate the long-tail problem, our motivation is to address this issue in \textbf{both two domains} which is novel and more challenging.
Table \ref{tab:summary} summarizes the difference between the proposed and prior methods. 
The contributions of the paper are four-fold. 
\begin{itemize}
    \item \textit{Firstly}, a \textit{\textbf{new metric is formulated for domain adaptation in long-tail semantic segmentation}}. Via the proposed metric, several constraints are introduced to the learning process including (1) \textit{multiple predictions per image} (i.e., one class per pixel) with class-balanced constraint; (2) \textit{local and global constraints via structural learning} with the novel conditional maximum likelihood loss. 
    \item \textit{Secondly}, \textit{\textbf{a Bijective Maximum Likelihood (BiMaL) loss}} is formed using a Maximum-likelihood formulation to model the global structure of a segmentation input and a Bijective function to map that segmentation structure to a deep latent space. The proposed BiMaL loss can be used with an unsupervised deep neural network to generalize on target domains.
    \item \textit{Thirdly}, a new \textit{\textbf{Conditional Maximum Likelihood (CoMaL) approach}} is introduced in the self-supervised domain adaptation framework to tackle the long-tail semantic scene segmentation problem. In this framework, the assumptions of pixel independence made by prior semantic segmentation methods are relaxed while more structural dependencies are efficiently taken into account.
    \item \textit{Finally}, the proposed BiMaL and CoMaL methods are evaluated on two popular large-scale visual semantic scene segmentation adaptation benchmarks, i.e., SYNTHIA $to$ Cityscapes and GTA $to$ Cityscapes, and \textit{\textbf{outperforms the prior segmentation methods by a large margin}}. 
\end{itemize}

Figure \ref{fig:first_fig} illustrates our state-of-the-art (SOTA) performance compared to prior approaches.

\begin{table*}[t]
\centering
\caption{ \textbf{Comparison in properties between our approaches and others}. Convolutional Neural Network (CNN), Generative Adversarial Net (GAN), Bijective Network (BiN), Multi-head Attention Network (MHA), Entropy Minimization (EntMin), Segmentation Map (Seg), Depth Map (Depth); $\ell_{IW}$: Image-wise Weighting Loss, $\ell_{CE}$: Cross-entropy Loss, $\ell_{Focal}$: Focal Loss $\ell_{adv}$: Adversarial Loss, $\ell_{Huber}$: Huber Loss. $\ell_{square}$: Maximum Squares Loss,  $\ell_{density}$: Maximum Likelihood Loss, $\ell_{CoMal}$: Conditional Maximum Likelihood Loss, $\ell_{Seesaw}$: Seesaw Loss, $\ell_{DropLoss}$: Drop Loss}%
\begin{tabular}{|c|c|c|c|c|c|c|c|}
\hline
\textbf{Methods} & \begin{tabular}{@{}c@{}} \textbf{Long-Tail}\\\textbf{Aware}  \end{tabular}  & \begin{tabular}{@{}c@{}} \textbf{Long-Tail} \\ {\textbf{Level}}\end{tabular}           & \begin{tabular}{@{}c@{}}\textbf{Structural}\\\textbf{Learning} \end{tabular} & \begin{tabular}{@{}c@{}}\textbf{Source} \\ \textbf{Label}\end{tabular} & \begin{tabular}{@{}c@{}} {\textbf{Target Domain}} \\ {\textbf{Transfer}} \end{tabular} & \textbf{Architecture}& \textbf{Designed Loss}  \\ 
\hline
AdaptPatch \cite{tsai2019domain} & \xmark& $-$& Weak (Binary label) & Seg  & \cmark & CNN+GAN & $\ell_{adv}$ \\
CBST \cite{zou2018unsupervised} &\xmark &$-$ & $-$ & Seg & \cmark & CNN & $\ell_{CE}$\\
ADVENT \cite{vu2019advent} &\xmark &$-$ & Weak (Binary label) & Seg  & \cmark & CNN+GAN & $\operatorname{EntMin} + \ell_{adv}$ \\
IntraDA \cite{pan2020unsupervised} & \xmark&$-$ & Weak (Binary label) & Seg & \cmark & CNN+GAN & $\operatorname{EntMin} + \ell_{adv}$ \\
\hline
Seesaw \cite{wang2021seesaw} & \cmark & Instance & $-$ & $-$ & \xmark &  CNN & $\ell_{Seesaw}$ \\
DropLoss \cite{DBLP:conf/aaai/Ting21} & \cmark & Instance & $-$ & $-$ & \xmark &  CNN & $\ell_{DropLoss}$ \\
\hline
SPIGAN \cite{lee2018spigan}  &\xmark &$-$ & Weak (Binary label) & Seg + Depth & \cmark & CNN+GAN & $\ell_{adv}$ \\
DADA \cite{vu2019dada} &\xmark &$-$ & Depth-aware Label & Seg + Depth & \cmark & CNN+GAN & $\ell_{adv}+\ell_{Huber}$ \\
\hline 
\hline
MaxSquare \cite{chen2019domain} & \begin{tabular}{@{}c@{}} Partial \\ \end{tabular} & Semantic & \begin{tabular}{@{}c@{}} Weak (Binary label) \end{tabular} & Seg & \cmark & CNN + GAN & $\ell_{square} + \ell_{IW}$\\
SAC \cite{Araslanov:2021:DASAC} & \cmark & Semantic & $-$ & Seg & \cmark & CNN & $\ell_{CE} + \ell_{Focal}$\\
\hline \hline
\textbf{BiMaL}         & \xmark  & $-$         & \textbf{Maximum Likelihood}   & Seg    & \cmark        & \begin{tabular}{@{}c@{}} \textbf{CNN + BiN} \end{tabular} & $\ell_{density}$ \\
\textbf{CoMaL} & \cmark & \begin{tabular}{@{}c@{}}\textbf{Semantic} \end{tabular}    & \begin{tabular}{@{}c@{}} \textbf{Conditional} \\ \textbf{Maximum Likelihood} \end{tabular}   &   Seg  & \cmark  & \begin{tabular}{@{}c@{}} \textbf{CNN + MHA} \end{tabular} &  $\ell_{CoMal}$ \\ 
\hline
\end{tabular}
\label{tab:summary}
\end{table*}

\section{Related Work}

Domain Adaptation has become one of the most popular research topics due to its ability to ease the notorious requirement of large amounts of labeled data in Computer Vision applications, especially those using deep learning methods. Domain discrepancy minimization \cite{ganin2015unsupervised, long2015learning, tzeng2017adversarial}, adversarial learning \cite{chen2018road,chen2017no, hoffman18a, hoffman2016fcns, hong2018CVPR, tsai2018learning}, entropy minimization \cite{murez2018CVPR, pan2020unsupervised, vu2019advent, zhu2017unpaired}, and self-training \cite{zou2018unsupervised} are the four primary approaches to Domain Adaptation.

\noindent
\textbf{Semantic Segmentation.} 
The performance provided by Fully Convolutional Networks in semantic segmentation applications makes them the most popular choice for the task, and the accuracy of FCNs improves further when incorporated  with an encoder-decoder structure.
Some of the first FCN applications \cite{long2015fully, chen2018deeplab} segmented images by incorporating spatial pooling after multiple convolutional layers. Works that followed \cite{lin2017refinenet, pohlen2017full} gathered more overall information and maintained accurate instance borders by combining upsampled, high-level feature maps with low-level feature maps prior to decoding.
Alternative approaches \cite{chen2018deeplab, DBLP:journals/corr/YuK15} have utilized dilated convolutions to improve the efficiency of models without sacrificing the field of view. Spatial pyramid pooling has been used by other recent works \cite{chen2017rethinking, chen2018encoder} to obtain contextual information at multiple levels. This approach grants more global information at higher layers in the network.
Deeplabv3+ \cite{chen2017rethinking} combined spatial pyramid pooling and the encoder-decoder structure in a new, efficient FCN architecture.
Recent works have utilized Transformer-based backbones \cite{xie2021segformer, daformer, transda} to learn a powerful semantic segmentation network.

\noindent
\textbf{Adversarial Training Methods.}
The approaches in this category are the preferred approaches of utilizing domain adaptation in semantic scene segmentation. 
The supervised segmentation training on the source domain and the adversarial training are designed in parallel.
The first GAN-based approach to semantic segmentation that utilized SDA was introduced by Hoffman et. al. \cite{hoffman2016fcns}. 
Chen et. al. \cite{chen2017no} made use of pseudo labels in conjunction with global and class-wise adaptation to improve their adversarial learning results.
Chen et. al. \cite{chen2018road} integrated target-guided distillation loss with a spatial-aware model to learn the real image style and spatial structures of urban scenes.
Hong et. al. \cite{hong2018CVPR} trained a conditional generator to reproduce features in real images to minimize the gap between these two domains. 
Tsai et. al. \cite{tsai2018learning} focused on the shared structures between these two domains, using adversarial training to predict similar label distributions across the source and target domains.
Other methods have investigated the efficacy of using generative networks to translate from a conditioned source to a novel target \cite{zhu2017unpaired, murez2018CVPR}. 
Hoffman et. al. \cite{hoffman18a} developed a Cycle-Consistent Adversarial network to adapt representations at the pixel and feature level.
Zhu et. al. \cite{zhu2018ECCV} presented a conservative loss function to promote moderate source examples, while avoiding the extreme examples during adversarial training.
Wu et al. \cite{wu2018dcan} utilized channel-wise alignment during generation and segmentation to further reduce domain shift.
Sakardis et. al. \cite{SDHV18} introduced a framework to segment foggy city images by gradually training on light, synthetic fog before moving to dense, real fog. 
Vapnik et. al. \cite{vapnik2009new} was the first to demonstrate the benefits of privileged information over traditional learning methods, granting additional data during training. 
Privileged information has been widely used since Vapnik's discovery \cite{hoffman2016learning, lopez2015unifying, mordan2018revisiting, Sharmanska_2013_ICCV} to improve model results.  
SPIGAN \cite{lee2018spigan} made use of privileged information to train a SDA model for semantic segmentation. Vu et. al.\cite{vu2019dada} created a depth-aware SDA framework to use privileged depth information, similar to SPIGAN.

\noindent
\textbf{Entropy Minimization Methods.} These approaches 
have become the preferred approach for semi-supervised learning \cite{grandvalet2005semi, springenberg2015unsupervised}. Vu et al. \cite{vu2019advent} pioneered the use of entropy minimization in semantic segmentation with domain adaptation: adversarial learning further improves the results of the minimization process. Using the entropy level of prediction, \cite{pan2020unsupervised, 10.1145/3474085.3475174} created an intradomain adaptation approach with two learning phases. The first phase performs the standard adaptation from the source to target domain, while the second aligns the easy and hard split in the target domain. 
Truong et al. \cite{truong2021bimal} introduced a novel bijective maximum likelihood approach to domain adaptation, which is a general form of entropy minimization to learn the global structure of the segmentation.
Self-training is another recently developed domain adaptation method for segmentation \cite{zou2018unsupervised, Araslanov:2021:DASAC} and classification \cite{NEURIPS2019_bf25356f}. 
In self-training, a new model is trained on unlabeled data by using pseudolabels derived from predictions of a trained model. 

\begin{figure*}[t]
    \centering
    \includegraphics[width=0.95\textwidth]{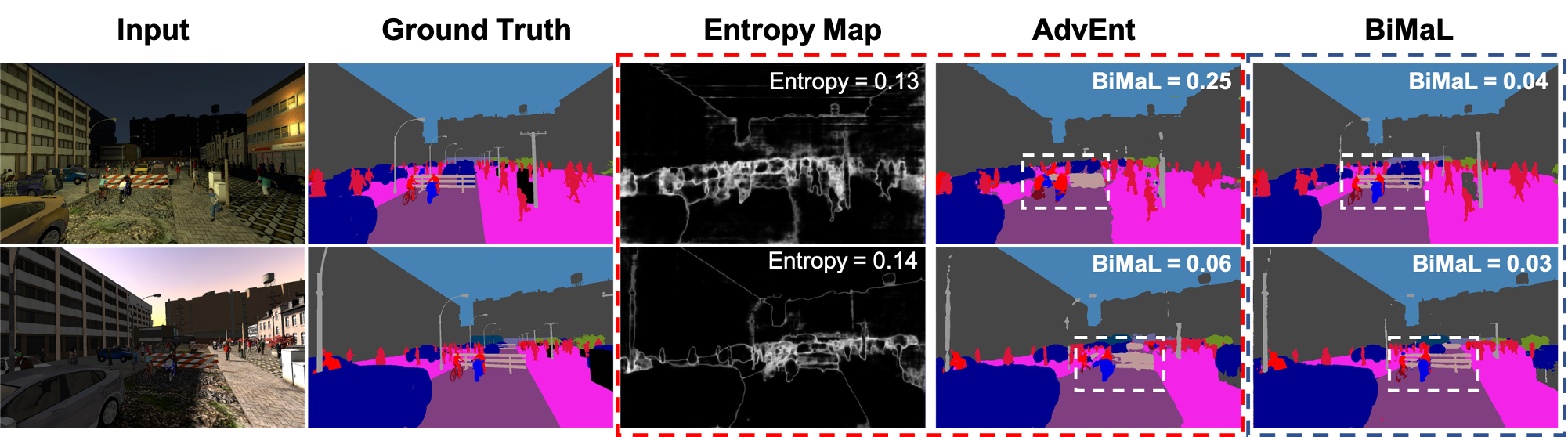}
    \caption{\textbf{Two images have the same entropy but one has a poor prediction (a top image) and one has an better prediction (a bottom image).} Columns 1 and 2 are an input image and a ground truth. Columns 3 and 4 are an entropy map and a prediction of AdvEnt \cite{vu2019advent}. Column 5 is the results of our proposed method. The two predictions produced by AdvEnt have similar entropy scores ($0.13$ and $0.14$). Meanwhile, the BiMaL value of the bottom prediction ($0.06$) is smaller than the top prediction ($0.14$). Our results in the last column, which have better BiMaL values than AdvEnt, can well model the structure of an image. In particular, our results have sharper results of a barrier and a rider (white dash box), and a clear boundary between road and sidewalk.}
    \label{fig:same_entropy_1}
\end{figure*}

\noindent
\textbf{Long-tail Recognition.} 
Ziwei et al.\cite{liu2019largescale} presented Open Long-tail Recognition to handle imbalanced classification, few-shot learning, and open-set recognition. 
They developed an integrated algorithm that transforms an image to a feature space, and their dynamic meta-embedding combines a direct image feature and a related memory feature, where the feature norm displays the familiarity of known classes. 
Jiawei et al.\cite{ren2020balanced} introduced Balanced Softmax, an unbiased extension of Softmax, to facilitate the distribution shift of labels between training and testing. Moreover, they introduced Balanced Meta-Softmax to enhance the long-tail learning by applying a complementary meta-sampler to estimate the optimal class sample rate.
Wang et al. \cite{wang2021seesaw} introduced seesaw loss to rebalance the fairness of gradients produced by positive and negative samples of a class with two regularization terms, i.e. mitigation and compensation.

\section{Cross-Domain Adaptation For Semantic Segmentation}

Let $\mathbf{x}_s \in \mathcal{X}_s \subset \mathbb{R}^{H \times W \times 3}$ be an input image of the source domain ($H$ and $W$ are the height and width of an image),  $\mathbf{x}_t \in \mathcal{X}_t \subset \mathbb{R}^{H \times W \times 3}$ be an input image of the target domain, $F: \mathcal{X} \to \mathcal {Y}$ where $\mathcal{X} = \mathcal{X}_s \cup \mathcal{X}_t$ be a semantic segmentation function 
that maps an input image to its corresponding segmentation map $\mathbf{y} \subset \mathbb{R}^{H \times W \times C}$, i.e. $\mathbf{y} = F(\mathbf{x}, \theta)$ ($C$ is the number of semantic classes). 
In general, given $N_s$ labeled training samples from a source domain 
$\mathcal{D}_s = \{\mathbf{x}_s^i, \hat{\mathbf{y}}_s^i\}_1^{N_s}$ 
and $N_t$ unlabeled samples from a target domain $\mathcal{D}_t = \{\mathbf{x}_t^i\}_1^{N_t}$, the 
unsupervised domain adaptation for semantic segmentation is formulated as: 
\begin{equation} \label{eqn:objective}
\small
\begin{split}
    \theta^{*} &= \arg\min_{\theta} \sum_{i,j} \big[\mathcal{L}_{s}(F(\mathbf{x}^i_s, \theta), \hat{\mathbf{y}}^i_s) + \mathcal{L}_{t}(F(\mathbf{x}^j_t, \theta))\big]\\
    &=\arg\min_{\theta} \Big[\mathbb{E}_{\mathbf{x}_s \sim p(\mathbf{x}_s), \hat{\mathbf{y}}_s \sim p(\hat{\mathbf{y}}_s)} \big[\mathcal{L}_{s}(F(\mathbf{x}_s, \theta), \hat{\mathbf{y}}_s)] \\
    &\quad \quad \quad \quad + \mathbb{E}_{\mathbf{x}_t \sim p(\mathbf{x}_t)} [\mathcal{L}_{t}(F(\mathbf{x}_t, \theta))\big]\Big]\\
    &=\arg\min_{\theta} \Big[\mathbb{E}_{\mathbf{y}_s \sim p(\mathbf{y}_s), \hat{\mathbf{y}}_s \sim p(\hat{\mathbf{y}}_s)} \big[\mathcal{L}_{s}(\mathbf{y}_s, \hat{\mathbf{y}}_s)] \\
    &\quad \quad \quad \quad + \mathbb{E}_{\mathbf{y}_t \sim p(\mathbf{y}_t)} [\mathcal{L}_{t}(\mathbf{y}_t)\big]\Big]\\
\end{split}
\end{equation}
where $\theta$ is the parameters of $F$, $p(\mathbf{\cdot})$ is the probability density function. As the labels for $\mathcal{D}_s$ are available, $\mathcal{L}_s$ can be efficiently formulated as a supervised cross-entropy loss:
\begin{equation}
    \small
    \mathcal{L}_{s}(\mathbf{y}_s, \hat{\mathbf{y}}_s) = -
    \sum_{h,w, c} \hat{\mathbf{y}}^{h,w,c}_s\log\left(\mathbf{y}^{h,w,c}_s\right)
    \label{Eqn2_crossentropy}
\end{equation}
where $\mathbf{y}^{h,w,c}$ and $\hat{\mathbf{y}}^{h,w,c}$ represent the predicted and ground-truth probabilities of the pixel at the location of $(h, w)$  taking the label of $c$, respectively. 
Meanwhile, $\mathcal{L}_t$ handles unlabeled data from
the target domain where the ground-truth labels are not available. To alleviate this label lacking issue, several forms of $\mathcal{L}_{t}(\mathbf{y}_t)$ have been exploited such as cross-entropy loss with pseudo-labels \cite{zou2018unsupervised}, Probability Distribution Divergence (i.e. Adversarial loss defined via an additional Discriminator) \cite{tsai2018learning, tsai2019domain}, %
or entropy formulation \cite{vu2019advent, pan2020unsupervised}.  

\noindent
\textbf{Entropy minimization revisited.}
By adopting the \mbox{Shannon} entropy formulation to the target prediction and constraining function $F$ to produce a high-confident prediction,
$\mathcal{L}_{t}$ can be formulated as %
\begin{equation} \label{eqn:entropy_loss}
\small
    \mathcal{L}_{t}(\mathbf{y}_t) = 
    \frac{-1}{\log(C)}\sum_{h,w,c}\mathbf{y}^{h,w,c}_t\log\left(\mathbf{y}^{h,w,c}_t\right).
\end{equation}
Although this form of $\mathcal{L}_{t}$ can give a direct assessment of the predicted segmentation maps, 
it tends to be dominated 
by the high probability areas (since the high probability areas 
produce a higher value 
updated gradient due to  $\lim_{\mathbf{y}^{h,w,c}_t \to 1}\frac{-\partial \mathcal{L}_t(\mathbf{y}_t)}{\partial \mathbf{y}^{h,w,c}_t} = \frac{1}{\log(C)}$ and $\lim_{\mathbf{y}^{h,w,c}_t \to 0}\frac{-\partial \mathcal{L}_t(\mathbf{y}_t)}{\partial \mathbf{y}^{h,w,c}_t} = -\infty$),  %
i.e. easy classes, rather than difficult classes \cite{vu2019advent}. 
More importantly, this is essentially a pixel-wise formation, where pixels are treated independently of each other.
Consequently, the structural information is usually neglected in this form. This issue could lead to a confusion point during training process where two predicted segmentation maps have similar entropy but different segmentation accuracy, one correct and other incorrect as shown in Fig \ref{fig:same_entropy_1}.

\begin{figure*}[t]
    \centering
    \includegraphics[width=0.8\textwidth]{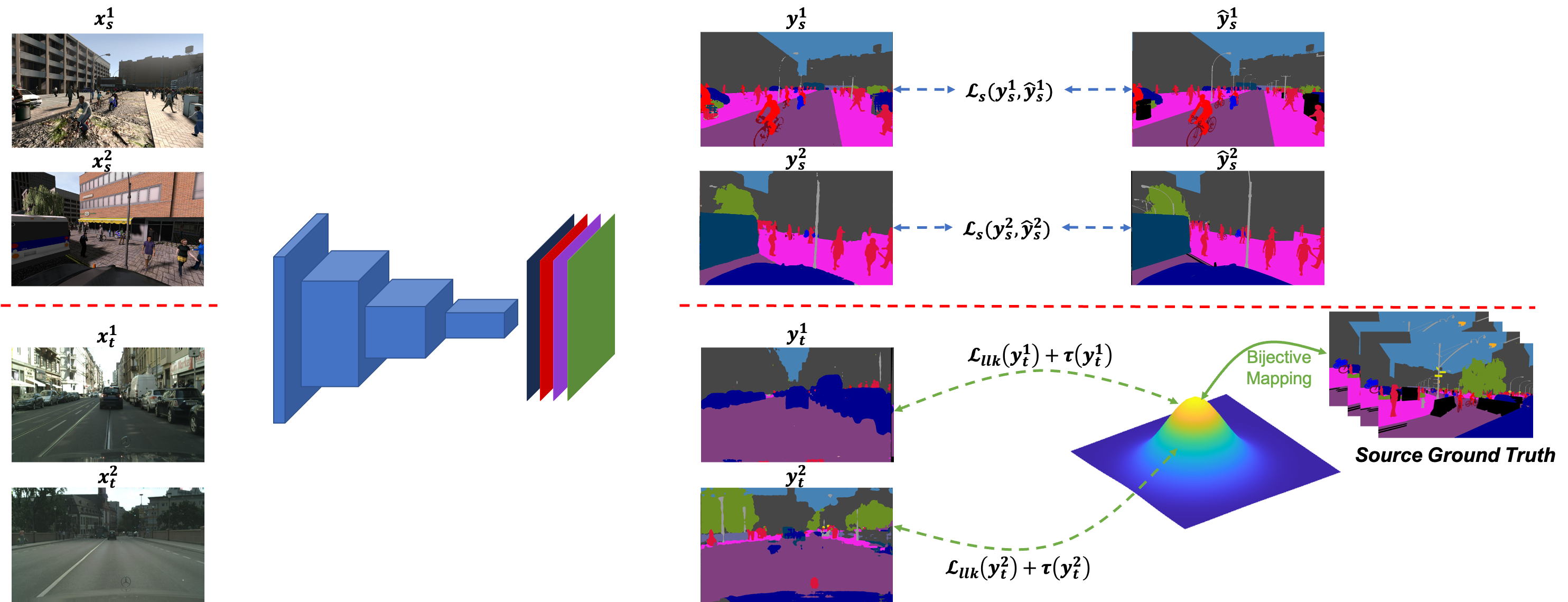}
    \caption{\textbf{The Proposed BiMaL Framework.} The RGB image input is firstly forwarded to a deep semantic segmentation network to produce a segmentation map. The supervised loss is employed on the source training samples. Meanwhile, the predicted segmentation on target training samples will be mapped to the latent space to compute the Bijective Maximum Likelihood loss. The bijective mapping network is trained on the ground-truth images of the source domain.}
    \label{fig:proposed_framework}
\end{figure*}

\subsection{Unaligned Domain Scores (UDS)}

In the entropy formulation, the pixel independent constraints are employed to convert the image-level metric to pixel-level metric. In contrast,
we propose an image-level UDS metric that can directly evaluate the structural quality of $\mathbf{y}_t$. 
Particularly, let $p_t(\mathbf{y}_t)$ and $q_t(\mathbf{y}_t)$ be the probability mass functions of the predicted distribution and the real (actual) distribution of the predicted segmentation map $\mathbf{y}_t$, respectively.
UDS metric measuring the efficiency of function $G$ on the target dataset can be expressed as follows:
\begin{equation}
\begin{split}
    \operatorname{UDS}%
    &= \mathbb{E}_{\mathbf{y}_t \sim p(\mathbf{y}_t)} \mathcal{L_Y}\left(p_t(\mathbf{y}_t),q_t(\mathbf{y}_t)\right) \\
    &= \int \mathcal{L_Y}\left(p_t(\mathbf{y}_t),q_t(\mathbf{y}_t)\right) p_t(\mathbf{y}_t) d\mathbf{y}_t \;,
\end{split}
    \label{eqn:UDSEqn}
\end{equation}
where 
$\mathcal{L_Y}\left(p_t(\mathbf{y}_t),q_t(\mathbf{y}_t)\right)$ defines the distance between %
two distributions $p_t(\mathbf{y}_t)$ and $q_t(\mathbf{y}_t)$. Since there is no label for sample in the target domain, the direct access to $q_t(\mathbf{y}_t)$ is not available. 
Note that although $\mathbf{x}_s$ and $\mathbf{x}_t$ could vary significantly in image space (e.g. difference in pixel appearance due to lighting, scenes, weather), their segmentation maps $\mathbf{y}_t$ and $\mathbf{y}_s$ share similar distributions in terms of both class distributions as well as global and local structural constraints (sky has to be above roads, trees should be on sidewalks, vehicles should be on roads, etc.). 
Therefore, one can practically adopted the prior knowledge learned from segmentation labels of the source domains for $q_t(\mathbf{y}_t)$ as
\begin{equation} \label{eqn:DDS_score_label_extend}
\begin{split}
    \operatorname{UDS} %
    &\approx \int \mathcal{L_Y}\left(p_t(\mathbf{y}_t),q_s(\mathbf{y}_t)\right) p_t(\mathbf{y}_t) d\mathbf{y}_t \;,
\end{split}
\end{equation}
where the distribution $q_s(\mathbf{y}_t)$ 
is the probability mass functions of the real distribution learned from ground-truth segmentation maps of $\mathcal{D}_s$. %
As a result, the proposed USD metric can be computed without the requirement of labeled target data for learning the density of segmentation maps in target domain. 
There are several choices for $\mathcal{L_Y}$ to estimate the divergence between the two distributions $p_t(\mathbf{y}_t)$ and $q_s(\mathbf{y}_t)$. In this paper, we adopt the common metric such as  Kullback–Leibler (KL) formula for $\mathcal{L_Y}$. 
Note that other metrics are also applicable in the proposed UDS formulation.
Moreover, to enhance the smoothness of the predicted semantic segmentation, a regularization term $\tau$ is imposed into $\mathcal{L_Y}$ as
\begin{equation} \label{eqn:LYUpdated}
\begin{split}
\small
    \mathcal{L_Y}\left(p_t(\mathbf{y}_t),q_s(\mathbf{y}_t)\right) &= \log\left(\frac{p_t(\mathbf{y}_t)}{q_s(\mathbf{y}_t)}\right) + \tau(\mathbf{y}_t). \\
\end{split}
\end{equation}
By computing UDS, one can measure the quality of the predicted segmentation maps $\mathbf{y}_t$ on the target data. 

In the next sections, we firstly discuss in details the learning process of $q_s(\mathbf{y}_t)$, and then derivations of the UDS metric for the novel Bijective Maximum Likelihood loss.

\subsection{Learning Distribution with Bijective Mapping on the Source Domain} \label{sec:bijective}

Let 
$G: \mathcal{Y} \to \mathcal{Z}$ 
be the bijective mapping function that maps a segmentation $\hat{\mathbf{y}}_s \in \mathcal{Y}$ to the latent space $\mathcal{Z}$, i.e. $\hat{\mathbf{z}}_s = G(\hat{\mathbf{y}}_s, \theta_F)$, where $\hat{\mathbf{z}}_s \sim q_z(\hat{\mathbf{z}}_s)$ is the latent variable, and $q_z$ is the prior distribution.
Then, the probability distribution $q_s(\hat{\mathbf{y}}_s)$ can be formulated via the change of variable formula:
\begin{equation} \label{eqn:bijective_mapping}
\small
    \log(q_s(\hat{\mathbf{y}}_s)) = \log\left(q_{z}(\hat{\mathbf{z}}_s)\right) + \log\left(\left|\frac{\partial G(\hat{\mathbf{y}}_s, \theta_G)}{\partial \hat{\mathbf{y}}_s}\right|\right),
\end{equation}
where $\theta_F$ is the parameters of $G$, $\left|\frac{\partial G(\hat{\mathbf{y}}_s, \theta_G)}{\partial \hat{\mathbf{y}}_s}\right|$ denotes the Jacobian determinant of function $G(\hat{\mathbf{y}}_s, \theta_G)$ with respect to $\hat{\mathbf{y}}_s$. 
To learn the mapping function, the negative log-likelihood
will be minimized as follows:
\begin{equation} \label{eqn:BijectiveLearning}
\footnotesize
\begin{split}
    \theta_G^{*} =& \arg\min_{\theta_G} %
    \mathbb{E}_{\hat{\mathbf{y}}_s \sim q_s(\hat{\mathbf{y}}_s)} \Big[-\log(q_s(\hat{\mathbf{y}}_s))\Big] \\
    =& \arg\min_{\theta_G} \mathbb{E}_{\hat{\mathbf{z}}_s \sim q_z(\hat{\mathbf{z}}_s)} \left[-\log\left(q_{z}(\hat{\mathbf{z}}_s)\right) - \log\left(\left|\frac{\partial G(\hat{\mathbf{y}}_s, \theta_G)}{\partial \hat{\mathbf{y}}_s}\right|\right)\right].
    \raisetag{40pt}
\end{split}
\end{equation}
In general, there are various choices for the prior distribution $q_z$. However, the ideal distribution should satisfy two 
criteria: (1) simplicity in the density estimation, and (2) easy in sampling. 
Considering the two criteria, we choose Normal distribution %
as the prior distribution $q_z$. 
Note that any other distribution is also feasible as long as it satisfies the mentioned criteria.

To enforce the information flow from a segmentation domain to a latent space with different abstraction levels, the bijective function $G$ can be further formulated as a composition of several sub-bijective functions $g_i$ as $G = g_1 \circ g_2 \circ ... \circ g_K$, 
where $K$ is the number of sub-functions. The Jacobian $\frac{\partial G}{\partial \mathbf{y}_s}$ can be derived by $\frac{\partial G}{\partial \hat{\mathbf{y}}_s} = \frac{\partial g_1}{\partial \hat{\mathbf{y}}_s} \cdot \frac{\partial g_2}{ \partial f_1} \cdots \frac{\partial g_K}{ \partial g_{K-1}}$. With this structure, the properties of each $g_i$ will define the properties for the whole bijective mapping function $G$. 
Interestingly, with this form, $G$ becomes a DNN structure when $g_i$ is a non-linear function built from a composition of convolutional layers. Several DNN structures \cite{dinh2015nice, dinh2017density,Duong_2017_ICCV, duong2020vec2face, glow, duong2019learning, truong2021fastflow} can be adopted for sub-functions.

\subsection{Bijective Maximum Likelihood (BiMAL) Loss} %
In this section, we present the proposed Bijective Maximum Likelihood (BiMaL) which can be used as the loss of target domain $\mathcal{L}_t$. 
\subsubsection{BiMaL formulation}
From Eqns. \eqref{eqn:DDS_score_label_extend} and \eqref{eqn:LYUpdated}, %
UDS metric can be rewritten as follows: %
\begin{equation} \label{eqn:KL_to_llk}
\small
\begin{split}
        \text{UDS} &= \int \left[\log\left(\frac{p_t(\mathbf{y}_t)}{q_s(\mathbf{y}_t)}\right)+  \tau(\mathbf{y}_t)\right]p_t(\mathbf{y}_t) d\mathbf{y}_t \\
        & = {\mathbb{E}}_{\mathbf{y}_t \sim p_t(\mathbf{y}_t)}\left[
        \log(p_t(\mathbf{y}_t))\right] \\
        & \quad - {\mathbb{E}}_{\mathbf{y}_t \sim p_t(\mathbf{y}_t)} \left[\log(q_s(\mathbf{y}_t)) \right] 
        +  {\mathbb{E}}_{\mathbf{y}_t \sim p_t(\mathbf{y}_t)}\left[\tau(\mathbf{y}_t)\right] \\
        &\leq {\mathbb{E}}_{\mathbf{y}_t \sim p_t(\mathbf{y}_t)} \left[-\log(q_s(\mathbf{y}_t)) + \tau(\mathbf{y}_t)\right]  
\end{split}    
\end{equation}

It should be noticed that with any form of the distribution $p_t$, the above inequality still holds as $p_t(\mathbf{y}_t) \in [0,1]$ and $\log(p_t(\mathbf{y}_t)) \leq 0$. Now, we define our Bijective Maximum Likelihood Loss as
\begin{equation} \label{eqn:BiMaL}
\begin{split}
    \mathcal{L}_t(\mathbf{y}_t) = -\log(q_s(\mathbf{y}_t))+ \tau(\mathbf{y}_t),
\end{split}
\end{equation}
where $\log(q_s(\mathbf{y}_t))$ defines the log-likelihood of $\mathbf{y}_t$ with respect to the density function $q_s$.
Then, by adopting the bijectve function $G$ learned from Eqn. \eqref{eqn:BijectiveLearning} using samples from source domain and the prior distribution $q_z$, the first term of $\mathcal{L}_t(\mathbf{y}_t)$ in Eqn. \eqref{eqn:BiMaL} can be efficiently computed via log-likelihood formulation:
\begin{equation} \label{eqn:define_llk}
\small
\begin{split}
    \mathcal{L}_{llk}(\mathbf{y}_t) &= -\log(q_s(\mathbf{y}_t)) \\
    &=  -\log\left(q_{z}(\mathbf{z}_t)\right) - \log\left(\left|\frac{\partial G(\mathbf{y}_t, \theta_G)}{\partial \mathbf{y}_t}\right|\right), 
\end{split}
\end{equation}
where $\mathbf{z}_t = G(\mathbf{y}_t, \theta_F)$. Thanks to the bijective property of the mapping function $G$, the minimum negative log-likelihood loss $\mathcal{L}_{llk}(\mathbf{y}_t)$
can be effectively computed via 
the density of the prior distribution $q_z$ and its associated Jacobian determinant $\left|\frac{\partial G(\mathbf{y}_t, \theta_G)}{\partial \mathbf{y}_t}\right|$. For the second term of $\mathcal{L}_t(\mathbf{y}_t)$, we further enhance the smoothness of the predicted semantic segmentation with the pair-wised formulation to encourage similar predictions for neighbourhood pixels with similar color:
\begin{equation} \label{eqn:define_tau}
\scriptsize
    \tau(\mathbf{y}_t) = \sum_{h,w}\sum_{h', w'} \exp \left\{-\frac{||\mathbf{x}_t^{h,w} -  \mathbf{x}_t^{h', w'}||_2^2}{2\sigma_1^2} - \frac{||\mathbf{y}_t^{h,w} - \mathbf{y}_t^{h', w'}||_2^2}{2\sigma_2^2}\right\}%
\end{equation}
where $(h', w') \in \mathcal{N}_{h, w}$ denotes the neighbourhood pixels of $(h, w)$, $\mathbf{x}^{h,w}$ represents the color at pixel $(h, w)$; and $\{\sigma_1, \sigma_2\}$ are the hyper parameters controlling the scale of Gaussian kernels. 
It should be noted that any regularizers \cite{chen2018deeplab, duong2029cvpr_automatic} enhancing the smoothness of the segmentation results can also be adopted for $\tau$.   
Putting Eqns. \eqref{eqn:BiMaL}, \eqref{eqn:define_llk}, \eqref{eqn:define_tau} to Eqn \eqref{eqn:objective}, the objective function can be rewritten as: %
\begin{equation}
\small
\begin{split}
         \theta^*=\arg\min_{\theta} \Big[&\mathbb{E}_{\mathbf{y}_s \sim p(\mathbf{y}_s), \hat{\mathbf{y}}_s \sim p(\hat{\mathbf{y}}_s)} \big[\mathcal{L}_{s}(\mathbf{y}_s, \hat{\mathbf{y}}_s)] \\ &+ \mathbb{E}_{\mathbf{y}_t \sim p(\mathbf{y}_t)} [\mathcal{L}_{llk}(\mathbf{y}_t) + \tau(\mathbf{y}_t)\big]\Big]
\end{split}
\end{equation}
Figure \ref{fig:proposed_framework} illustrates our proposed BiMaL framework to learn the deep segmentation network $G$. Also, we can prove that direct entropy minimization as Eqn. \eqref{eqn:entropy_loss} is just a particular case of our log likelihood maximization. We will further discuss how our maximum likelihood can cover the case of pixel-independent entropy minimization in Section \ref{sec:MLE_Entropy}.

\noindent
\subsubsection{BiMaL properties}  \label{sec:MLE_Entropy}

\textit{\textbf{Global Structure Learning.}} Sharing similar property with \cite{duong2016dam_cvpr, duong2019dam_ijcv, duong2020vec2face, Duong_2017_ICCV, 9108692}, from Eqn. \eqref{eqn:bijective_mapping}, as the learned density function is adopted for the entire segmentation map $\hat{\mathbf{y}}_s$, the global structure in $\hat{\mathbf{y}}_s$ can be efficiently captured and modeled.

\noindent
\textit{\textbf{Tractability and Invertibility.}} Thanks to the designed bijection F, the complex distribution of segmentation maps can be efficiently captured. Moreover, the mapping function is bijective, and, therefore, both inference and generation  are exact and tractable.
\subsubsection{Relation to Entropy Minimization} \label{sec:MLE_Entropy}
The first term of UDS in Eqn. \eqref{eqn:KL_to_llk} can be derived as
\begin{equation}
\small
    \begin{split}
        & \int \log\left(\frac{p_t(\mathbf{y})}{q_s(\mathbf{y})}\right)p_t(\mathbf{y}_t) d\mathbf{y}_t \geq 0 \\
        \Leftrightarrow & {\mathbb{E}}_{\mathbf{y}_t \sim p_t(\mathbf{y}_t)}\left[\log(p_t(\mathbf{y}_t)) - \log(q_s(\mathbf{y}_t))\right] \geq 0 \\
        \Leftrightarrow & {\mathbb{E}}_{\mathbf{y}_t \sim p_t(\mathbf{y}_t)}\left[-\log(q_s(\mathbf{y}_t))\right] \geq {\mathbb{E}}_{\mathbf{y}_t \sim p_t(\mathbf{y}_t)}\left[-\log(p_t(\mathbf{y}_t))\right]\\
        \Leftrightarrow & \mathbb{E}_{\mathbf{y}_t \sim p_t(\mathbf{y}_t)}[\mathcal{L}_{llk}(\mathbf{y}_t)] \geq \text{Ent}(\mathbf{Y}_t)%
        \raisetag{40pt}
    \end{split}
\end{equation}
where $\mathbf{Y}_t$ is the random variable with possible values $\mathbf{y}_t \sim p_t(\mathbf{y}_t)$, and $\text{Ent}(\mathbf{Y}_t)$ denotes the entropy of the random variable $\mathbf{Y}_t$.
It can be seen that the proposed negative log-likelihood $\mathcal{L}_{llk}$ is an upper bound of the entropy of $\mathbf{Y}_t$. Therefore, minimizing our proposed BiMaL loss will also enforce the entropy minimization process. 
Moreover, by not assuming pixel independence, our proposed BiMaL can model and evaluate structural information at the image-level better than previous pixel-level approaches \cite{chen2019domain, pan2020unsupervised, vu2019advent}.

\section{Cross-Domain Long-tail Adaptation For Semantic Segmentation}
As shown in Fig. \ref{fig:long_tailed_distribition}, 
in Semantic Segmentation task, there is another challenging issue that can significantly affect the performance of domain adaptation approaches, i.e. the long-tail issue. 
This occurs in both labeled source domain \textbf{and} new \textbf{\textit{unlabeled}} target domains. 
Unlike prior works where only labeled part is on their focus to alleviate the long-tail problem, our motivation is to address this issue in \textbf{both two domains} which is novel and more challenging.
In this section, we further investigate the long-tail issue for adaptation in semantic segmentation. Then, a novel Conditional
Maximum Likelihood (CoMaL) loss is further introduced to address the long-tail problem. Similar to BiMaL loss, the assumptions of pixel independence made by prior semantic segmentation methods are relaxed while more structural dependencies are efficiently taken into account during learning process.

\subsection{Long-tail Problem in Domain Adaptation} \label{sec:long-tail-issue}
Different from classification problem where
a single class is predicted for each input sample/image, in semantic segmentation, $\mathbf{y}_s$ and $\mathbf{y}_t$ are segmentation maps whose each pixel has its own predicted class. Let $N = H \times W$ be the total number of pixels in an input image where $H$ and $W$ are the height and width of that image.
In addition, let $y^i_s$ (or  $y^i_t$) be the predicted class of the $i^{th}$ pixel in $\mathbf{y}_s$ (or $\mathbf{y}_t$). Eqn. \eqref{eqn:objective} can be rewritten as follows.

\begin{equation} \label{AA}
\scriptsize
    \begin{split}
    \theta^* &=\arg\min_{\theta} \Bigg[\int \mathcal{L}_s(\mathbf{y}_s, \mathbf{\hat{y}}_s)q_s(\mathbf{y}_s) q_s(\mathbf{\hat{y}}_s)d\mathbf{y}_s d\mathbf{\hat{y}}_s 
    + \int \mathcal{L}_t(\mathbf{y}_t)p_t(\mathbf{y}_t)d\mathbf{y}_t\Bigg]\\
    &=\arg\min_{\theta} \Bigg[\int \sum_{i=1}^N\mathcal{L}_s(y^i_s, \hat{y}^i_s) q_s(y^i_s)q_s(\mathbf{y}^{\setminus i}_s | y^i_s)q_s(\mathbf{\hat{y}}_s)d\mathbf{y}_s d\mathbf{\hat{y}}_s  \\
    &\qquad\qquad\qquad\qquad\qquad\qquad\qquad  
    + \int \sum_{i=1}^N\mathcal{L}_t(y^i_t) p_t(y^i_t)p_t(\mathbf{y}^{\setminus i}_t | y^i_t)d\mathbf{y}_t\Bigg]
    \end{split}
\end{equation}
where $\mathbf{y}^{\setminus i}_s$ (or  $\mathbf{y}^{\setminus i}_t$) is a predicted segmentation map conditional on the prediction of the $i^{th}$ pixel;
$q_s(y^i_s)$ and $ p_t(y^i_t)$ 
are the class distributions received by a pixel of the data,
$q_s(\mathbf{y}^{\setminus i}_s | y^i_s)$ and $p_t(\mathbf{y}^{\setminus i}_t | y^i_t)$ 
represent the conditional structure constraints of the segmentation. %
From Eqn. \eqref{AA}, two main properties are required to be addressed during optimization process, i.e. (1) \textit{long-tail problem} and (2) \textit{structural learning} for semantic segmentation.

\noindent
\textit{\textbf{Long-tail Problem. }} In practice, the class distributions 
$q_s(y^i_s)$ and $ p_t(y^i_t)$ 
in both source and target domains 
suffer heavy long-tail problems as illustrated in Fig. \ref{fig:long_tailed_distribition}.
As the learning gradient is updated based on predicted class of each pixel, there is a bias toward the class occupying large regions. In other words, the gradients from pixels of the head classes (i.e. road, sky, sidewalk, cars) are largely dominant over the tail classes (i.e. poles, fences, or pedestrians) in an image. 
\begin{figure*}[!t]
    \centering
    \includegraphics[width=0.99\textwidth]{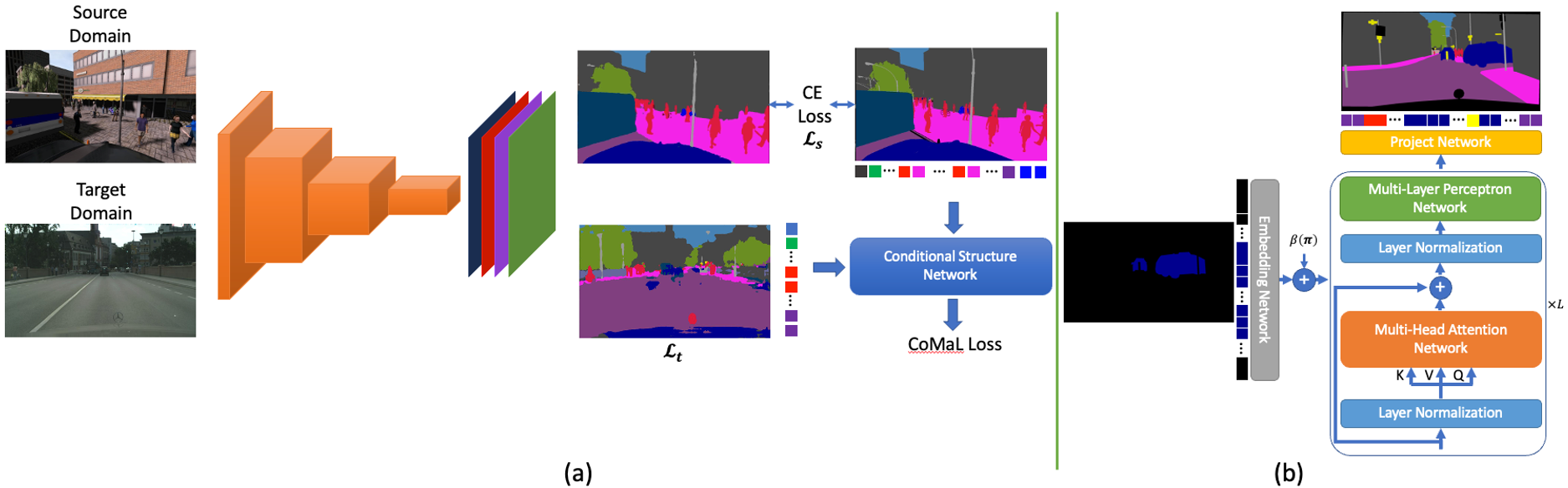}
    \caption{\textbf{The Proposed Framework}.
    (a) The Self-supervised Adaptation Framework. 
    The cross-entropy (CE) loss ($\mathcal{L}_s$) is applied on the source domain and the target loss $\mathcal{L}_t$ is deployed on the target domains. The segmentation on both domains are forwarded into the Conditional Structure Network to compute the Conditional Maximum Likelihood (CoMaL) loss. (b) The Conditional Structure Network.}
    \label{fig:framework}
\end{figure*}
For example, let the segmentation image have only two classes where 
$\mathbf{q}_s(y_s^i=0) \ll \mathbf{q}_s(y_s^i=1)$, 
$N =N_0 + N_1$ where $N_0$ and $N_1$ are the number of pixels of class $0$ and class $1$, respectively. In addition, 
when the number of pixels of class 1 largely dominants class 0, i.e. $N_0 \ll N_1$, an inequality can be derived as follows.
\begin{equation} \label{eqn:inequality}
\scriptsize
\begin{split}
    &\left|\left|\frac{\partial \int \sum_{i=1}^{N}\mathcal{L}_s(y^i_s, \hat{y}^i_s) q_s(y^i_s)q_s(\mathbf{y}^{\setminus i}_s | y^i_s )q_s(\mathbf{\hat{y}}_s)d\mathbf{y}_s d\mathbf{\hat{y}}_s}{\partial \mathbf{y}^{(0)}_{s}}\right|\right| \ll \\
    &\quad\quad \left|\left|\frac{\partial \int \sum_{i=1}^{N}\mathcal{L}_s(y^i_s, \hat{y}^i_s) q_s(y^i_s)q_s(\mathbf{y}^{\setminus i}_s | y^i_s )q_s(\mathbf{\hat{y}}_s)d\mathbf{y}_s d\mathbf{\hat{y}}_s}{\partial \mathbf{y}^{(1)}_{s}}\right|\right|
\end{split}
\end{equation}
where $||.||$ is the magnitude of the vector, $\mathbf{y}^{(0)}_s$ (or $\mathbf{y}^{(1)}_s$) represent the predicted probabilities of label 0 (or 1).
As shown in Eqn. \eqref{eqn:inequality}, when the distribution of class 1 largely dominants class 0, the losses defined in the source domain 
incline to produce large gradient updates for the head classes; meanwhile, the gradients of tail classes tend to be suppressed.
A similar observation is also made in \cite{wang2021seesaw}.
Similar to the source domain, this long-tail problem also occurs in the target domain.  %

\noindent
\textit{\textbf{Structural Learning for Semantic Segmentation in Source and Target domain. }} In Eqn. \eqref{AA}, one can see that both terms of 
$q_s(\mathbf{y}^{\setminus i}_s | y^i_s)$ and $p_t(\mathbf{y}^{\setminus i}_t | y^i_t)$ 
reflect the dependencies of pixels in their predictions. In prior works which focus only on learning semantic segmentation on source domain (i.e. where labels are available), supervised loss can help to partially and implicitly maintain the structure in $\mathbf{y}_s$. However, as labels are not available in the target domain, pixel independence is usually assumed for $\mathcal{L}_t$ \cite{vu2019advent, pan2020unsupervised, Araslanov:2021:DASAC, zhang2021prototypical} to be applicable. Then, only a weak indication of structural learning via Adversarial loss is adopted. 

In the next sections, we first propose a novel metric that addresses both properties in its formulation. Then its derivation as well as the conditional structural learning of 
$q_s(\mathbf{y}^{\setminus i}_s | y^i_s)$ and $p_t(\mathbf{y}^{\setminus i}_t | y^i_t)$ 
are presented for the learning process. 

\subsection{Learning Segmentation From Distributions with Conditional Maximum Likelihood (CoMaL) loss}

As in Section \ref{sec:long-tail-issue}, the long-tail issues occur in semantic segmentation due to the imbalance in class distributions, where the densities of head classes dominates those of the tail classes. Therefore, rather than directly deriving the formulation of Eqn. \eqref{AA} with the given training data distribution, we first assume a presence of an ideal dataset whose classes are uniformly distributed. Then, based on this ideal distribution, the derivation are obtained for all terms in Eqn. \eqref{AA}. Finally, the presence of the ideal data is removed for the training procedure to be applicable in practise. By this way, the long-tail issue can be efficiently addressed and practically adopted for a given training data. 

Formally, let $q'_s(\mathbf{y}_s), q'_s(\mathbf{\hat{y}}_s)$ be the ideal distributions of $\mathbf{y}_s$ and $\mathbf{\hat{y}}_s$ in source domain, and $p'_t(\mathbf{y}_t)$ represent the ideal distribution of the target domain. By learning from these ideal distributions, the numbers of pixels across classes in both domains become more balanced, and, therefore, alleviating the long-tail problem of semantic segmentation in predictions of both source and target domains.
Learning an SDA model with respect to $q'_s(\mathbf{y}_s), q'_s(\mathbf{\hat{y}}_s)$ and $p'_t(\mathbf{y}_t)$ can be adopted to Eqn. \eqref{AA} as follows.

\begin{equation} \label{eqn:optimization_data_viewpoint}
\scriptsize
\begin{split}
    \theta^* &= \arg\min_{\theta} \Bigg[\int \mathcal{L}_s(\mathbf{y}_s, \mathbf{\hat{y}}_s)q'_s(\mathbf{y}_s) q'_s(\mathbf{\hat{y}}_s)d\mathbf{y}_s d\mathbf{\hat{y}}_s  + \int \mathcal{L}_t(\mathbf{y}_t)p'_t(\mathbf{y}_t)d\mathbf{y}_t\Bigg] \\
    &= \arg\min_{\theta} \Bigg[\mathbb{E}_{\mathbf{y}_s \sim q_s(\mathbf{y}_s), \mathbf{\hat{y}_s} \sim q_s(\mathbf{\hat{y}}_s)} \mathcal{L}_s(\mathbf{y}_s, \mathbf{\hat{y}}_s)\frac{q'_s(\mathbf{y}_s) q'_s(\mathbf{\hat{y}}_s)}{q_s(\mathbf{y}_s)q_s( \mathbf{\hat{y}}_s)}  \\
    &\qquad\qquad\qquad\qquad\qquad\qquad\qquad\qquad\qquad  
    + \mathbb{E}_{\mathbf{y} \sim p_t(\mathbf{y}_t)} \mathcal{L}_t(\mathbf{y}_t)\frac{p'_t(\mathbf{y}_t)}{p_t(\mathbf{y}_t)}\Bigg] \\
\end{split}
\end{equation}
The fractions between the ideal distributions and data distributions, i.e. $\frac{q'_s(\mathbf{y}_s) q'_s(\mathbf{\hat{y}}_s)}{q_s(\mathbf{y}_s)q_s( \mathbf{\hat{y}}_s)}$ and $\frac{p'_t(\mathbf{y}_t)}{p_t(\mathbf{y}_t)}$, can be interpreted as the complement of the long-tail distributions to improve the class balanced during learning process, and, hence, enhance the robustness of the learned segmentation model against long-tail data. 

It is noticed that, 
the fraction between the ideal distribution $q'_s(\mathbf{\hat{y}}_s)$ and the long-tailed distribution $q_s(\mathbf{\hat{y}}_s)$ is a constant as these are the distributions over the ground truths. Therefore, it can be efficiently ignored during the optimization process.  
Eqn. \eqref{eqn:optimization_data_viewpoint} can be rewritten as follows:

\begin{equation} \label{eqn:optimization_without_likelihood}
\scriptsize
\begin{split}
    \theta^* &= \arg\min_{\theta} \Bigg[\mathbb{E}_{\mathbf{y}_s \sim q_s(\mathbf{y}_s), \mathbf{\hat{y}_s} \sim q_s(\mathbf{\hat{y}}_s)} \mathcal{L}_s(\mathbf{y}_s, \mathbf{\hat{y}}_s)\frac{q'_s(\mathbf{y}_s)}{q_s(\mathbf{y}_s)}  \\
    &\qquad\qquad\qquad\qquad\qquad\qquad\qquad\qquad\qquad
    + \mathbb{E}_{\mathbf{y} \sim p_t(\mathbf{y}_t)} \mathcal{L}_t(\mathbf{y}_t)\frac{p'_t(\mathbf{y}_t)}{p_t(\mathbf{y}_t}\Bigg] \\
    &= \arg\min_{\theta} \Bigg[\mathbb{E}_{\mathbf{y}_s \sim q_s(\mathbf{y}_s), \mathbf{\hat{y}_s} \sim q_s(\mathbf{\hat{y}}_s)} \sum_{i=1}^{N} \mathcal{L}_s(y^i_s, \hat{y}^i_s) \frac{q'_s(y^i_s)q'_s(\mathbf{y}^{\setminus i}_s|y^i_s)}{q_s(y^i_s)q_s(\mathbf{y}^{\setminus i}_s |y^i_s)}  \\
    &\qquad\qquad\qquad\qquad + \mathbb{E}_{\mathbf{y} \sim p_t(\mathbf{y}_t)} \sum_{i=1}^{N}\mathcal{L}_t(y^i_t) \frac{p'_t(y^i_t)p'_t(\mathbf{y}^{\setminus i}_t|y^i_t)}{p_t(y^i_t)p_t(\mathbf{y}^{\setminus i}_t |y^i_t)} \Bigg]
\end{split}
\end{equation}

Notice that as the labels of samples in this domain are not accessible, a direct computation of distributions in the target domain, i.e. $p_t(\mathbf{y}_t)$ and $p'_t(\mathbf{y}_t)$, is not feasible. 
Generally, one can see that although input images of source domain ($\mathbf{x}_s$) and target domain ($\mathbf{x}_t$) could vary significantly in the image space according to their differences in pixel appearance across domains (such as lighting, scenes, weather, etc.), their segmentation maps $\mathbf{y}_s$ and $\mathbf{y}_t$ share similar class distributions as well as global and local structure (sky has to be above roads, trees should be on side-walks, vehicles should be on roads, etc.). Therefore, the prior knowledge of segmentation maps $q_s(\mathbf{y}_s)$ and $q'_s(\mathbf{y}_s)$ can be practically adopted for $p_t(\mathbf{y}_t)$ and $p'_t(\mathbf{y}_t)$, respectively. 
As a result, Eqn. \eqref{eqn:optimization_without_likelihood} can be rewritten as follows:

\begin{equation} \label{eqn:optimization_with_assumption}
\scriptsize
\begin{split}
    \theta^* &= \arg\min_{\theta} \Bigg[\mathbb{E}_{\mathbf{y}_s \sim q_s(\mathbf{y}_s), \mathbf{\hat{y}_s} \sim q_s(\mathbf{\hat{y}}_s)} \sum_{i=1}^{N}\mathcal{L}_s(y^i_s, \hat{y}^i_s) \frac{q'_s(y^i_s)q'_s(\mathbf{y}^{\setminus i}_s|y^i_s)}{q_s(y^i_s)q_s(\mathbf{y}^{\setminus i}_s |y^i_s)}  \\
    &\qquad\qquad\qquad\qquad\qquad + \mathbb{E}_{\mathbf{y} \sim q_s(\mathbf{y}_t)} \sum_{i=1}^{N} \mathcal{L}_t(y^i_t) \frac{q'_s(y^i_t)q'_s(\mathbf{y}^{\setminus i}_t|y^i_t)}{q_s(y^i_t)q_s(\mathbf{y}^{\setminus i}_t |y^i_t)} \Bigg]
\end{split}
\end{equation}

By taking the logarithm of Eqn  \eqref{eqn:optimization_with_assumption}, a new objective function can be derived as follows (the full proof of Eqn. \eqref{eqn:take_log} will be available in the supplementary material):
\begin{equation} \label{eqn:take_log}
\footnotesize
\begin{split}
    &\theta^* \simeq \arg\min_{\theta} \Bigg[\mathbb{E}_{\mathbf{y}_s \sim q_s(\mathbf{y}_s), \mathbf{\hat{y}_s} \sim q_s(\mathbf{\hat{y}}_s)} \mathcal{L}_s(\mathbf{y}_s, \mathbf{\hat{y}}_s) + \mathbb{E}_{\mathbf{y} \sim q_s(\mathbf{y}_t))} \mathcal{L}_t(\mathbf{y}_t)\\
    &+\frac{1}{N}\sum_{i=1}^{N}
    \Bigg(\mathbb{E}_{\mathbf{y}_s \sim q_s(\mathbf{y}_s), \mathbf{\hat{y}_s} \sim q_s(\mathbf{\hat{y}}_s)} \log\left(\frac{q'_s(y^i_s)}{q_s(y^i_s)}\right) 
    \\
    &\quad\quad\quad\quad\quad\quad\quad\quad\quad\quad\quad\quad\quad
    +\mathbb{E}_{\mathbf{y} \sim q_s(\mathbf{y}_t)} \log\left(\frac{q'_s(y^i_t)}{q_s(y^i_t)}\right)\Bigg)\\
    &+\frac{1}{N}\sum_{i=1}^{N}\Bigg(\mathbb{E}_{\mathbf{y}_s \sim q_s(\mathbf{y}_s), \mathbf{\hat{y}_s} \sim q_s(\mathbf{\hat{y}}_s)} \log\left(\frac{q'_s(\mathbf{y}^{\setminus i}_s |y^i_s)}{q_s(\mathbf{y}^{\setminus i}_s |y^i_s)}\right)\\
    &\quad\quad\quad\quad\quad\quad\quad\quad\quad\quad\quad\quad\quad
    +\mathbb{E}_{\mathbf{y} \sim q_s(\mathbf{y}_t)} \log\left(\frac{q'_s(\mathbf{y}^{\setminus i}_t|y^i_t)}{q_s(\mathbf{y}^{\setminus i}_t |y^i_t)}\right)\Bigg)\Bigg]
\end{split}
\end{equation}

\noindent
From Eqn. \eqref{eqn:take_log}, several constraints are introduced for the learning process. 

\noindent
\textit{\textbf{Base Terms For Domain Adaptation}}. The first two terms of $\mathcal{L}_{s}$ and $\mathcal{L}_{t}$ are the supervised and self-supervised terms which help to embed the knowledge of labeled samples from source domain $\{ \mathcal{X}_s,  \mathcal{Y}_s\}$ and unlabeled samples of target domain $\{ \mathcal{X}_t\}$. %

\noindent
\textit{\textbf{Class-Balanced Learning}}.
The next two terms $\log\left(\frac{q'_s(\mathbf{y}^i_s)}{q_s(\mathbf{y}^i_s)}\right)$ and $\log\left(\frac{q'_s(\mathbf{y}^i_t)}{q_s(\mathbf{y}^i_t)}\right)$ 
can be computed from the training dataset during learning procedure. Notice that since $q'_s(\mathbf{y}^i_s)$ is a uniform distribution, its likelihood is a constant number. Therefore, this term can be efficiently ignored during learning process. 
Let us denote these fractions as the class-weighted loss $\mathcal{L}_{cls}$, e.g. $\mathcal{L}_{cls}(y^i_s) = \log\left(\frac{q'_s(y^i_s)}{q_s(y^i_s)}\right)$. 

\noindent
\textit{\textbf{Conditional Structural Learning}}.
The last two terms $\log\left(\frac{q'_s(\mathbf{y}^{\setminus i}_s |y^i_s)}{q_s(\mathbf{y}^{\setminus i}_s |y^i_s)}\right)$ and $\log\left(\frac{q'_s(\mathbf{y}^{\setminus i}_t|y^i_t)}{q_s(\mathbf{y}^{\setminus i}_t |y^i_t)}\right)$ regularize the conditional structure of the segmentation. 
In practice, as the ideal dataset is not available, the direct access to the conditional distributions $q'_s(\mathbf{y}^{\setminus i}_s |\mathbf{y}^i_s)$ and $q'_s(\mathbf{y}^{\setminus i}_t |\mathbf{y}^i_t)$ is not feasible. Therefore, rather than directly computing these terms, they can be derived as follows.

\begin{equation} \label{eqn:CoMaL}
\footnotesize
\begin{split}
    &\mathbb{E}_{\mathbf{y}_s \sim q_s(\mathbf{y}_s), \mathbf{y_s} \sim q_s(\mathbf{\hat{y}}_s)}\log\left(\frac{q'_s(\mathbf{y}^{\setminus i}_s |y^i_s)}{q_s(\mathbf{y}^{\setminus i}_s |y^i_s)}\right) + \mathbb{E}_{\mathbf{y} \sim q_s(\mathbf{y}_t)} \log\left(\frac{q'_s(\mathbf{y}^{\setminus i}_t|y^i_t)}{q_s(\mathbf{y}^{\setminus i}_t |y^i_t)}\right) \\
    &= \mathbb{E}_{\mathbf{y}_s \sim q_s(\mathbf{y}_s), \mathbf{y_s} \sim q_s(\mathbf{\hat{y}}_s)} \left[
    \log q'_s(\mathbf{y}^{\setminus i}_s |y^i_s) - \log q_s(\mathbf{y}^{\setminus i}_s |y^i_s)\right] \\
    &\quad\quad\quad\quad\quad\quad
    + \mathbb{E}_{\mathbf{y} \sim q_s(\mathbf{y}_t)}\left[ \log q'_s(\mathbf{y}^{\setminus i}_t|y^i_t) -\log q_s(\mathbf{y}^{\setminus i}_t |y^i_t)\right] \\
    &\leq \mathbb{E}_{\mathbf{y}_s \sim q_s(\mathbf{y}_s), \mathbf{y_s} \sim q_s(\mathbf{\hat{y}}_s)}  \underbrace{-\log q_s(\mathbf{y}^{\setminus i}_s |y^i_s)}_{\mathcal{L}_{CoMaL}(\mathbf{y}^{\setminus i}_s)} + \mathbb{E}_{\mathbf{y} \sim q_s(\mathbf{y}_t)} \underbrace{-\log q_s(\mathbf{y}^{\setminus i}_t |y^i_t)}_{\mathcal{L}_{CoMaL}(\mathbf{y}^{\setminus i}_t)}
\end{split}
\end{equation}
where $\mathcal{L}_{CoMaL}$ is our proposed Conditional Maximum Likelihood loss.
For any form of ideal distribution $q'_s(\cdot)$, our above inequality holds true as $q'_s(\cdot) \in [0, 1] \Leftrightarrow \log q'_s(\cdot) \leq 0$. Thanks to this inequality, the requirement of the ideal dataset can be effectively relaxed for the learning process. 

As a result. the entire self-supervised domain adaptation in long-tail segmentation framework can be optimized as:
\begin{equation}
\scriptsize
\begin{split}
    \theta^* = &\arg\min_{\theta}\frac{1}{N}\Bigg\{
    \mathbb{E}_{\mathbf{y}_s \sim q_s(\mathbf{y}_s), \mathbf{\hat{y}_s} \sim q_s(\mathbf{\hat{y}}_s)} \bigg[\mathcal{L}_{s}(\mathbf{y}_s, \mathbf{\hat{y}}_s) \\
    &\quad\quad\quad\quad\quad\quad\quad\;\; 
    + \frac{1}{N}\sum_{i=1}^N\Big(\mathcal{L}_{cls}(\hat{y}_s^i) + \mathcal{L}_{CoMaL}(\mathbf{\hat{y}}^{\setminus i}_s)\Big)\bigg] \\ 
     &+\mathbb{E}_{\mathbf{y}_t \sim q_s(\mathbf{y}_t)} \bigg[\mathcal{L}_{t}(\mathbf{y}_t) + \frac{1}{N}\sum_{i=1}^N\Big(\mathcal{L}_{cls}(y_t^i) + \mathcal{L}_{CoMaL}(\mathbf{y}^{\setminus i}_t)\Big)\bigg]
    \Bigg\}
\end{split}
\raisetag{50pt}
\end{equation}
In our approach, the loss $\mathcal{L}_{t}$ on the target domain is defined as the self-supervised loss with pseudo labels as \cite{Araslanov:2021:DASAC}. Fig. \ref{fig:framework}(a) illustrates our proposed framework.
In the next section,
we further discuss the learning process of the conditional structure $q_s(\mathbf{y}^{\setminus i}_s |y^i_s)$ on the ground truth of the source domain.

\subsection{The Conditional Structure Learning} \label{sec:ConditionalStructureLearning}

Given the segmentation dataset of the source domain, $\mathbf{y}_s \in \mathcal{Y}_s$, the structural condition $q_s(\mathbf{y}_s^{\setminus i} | y^i_s)$ can be auto-regressively modeled as follows:
\begin{equation} \label{eqn:regressive_with_order}
\footnotesize
\begin{split}
    \theta_G^* &= \arg\min_{\theta_G} -\mathbb{E}_{\mathbf{y}_s \sim \mathcal{Y}_s, \pi_i \in \Pi} \left[\log(q_s(\mathbf{y}_s^{\setminus i} | y^i_s))\right] \\
    &= \arg\min_{\theta_G} -\mathbb{E}_{\mathbf{y}_s \sim \mathcal{Y}_s, \pi_i \in \Pi} \sum_{j=2}^N \log(q_s(\mathbf{y}_s^{\pi^i_j} | \mathbf{y}^{\pi^i_{1}}_s..\mathbf{y}^{\pi^i_{j-1}}_s, \theta_G, \pi^i))
\end{split}
\raisetag{40pt}
\end{equation}
where $\pi^i$ is a permutation of $\{1..N\}$ and $\pi^i_1 = i$, $\theta_G$ is the parameters of the generative model $G$. 
The learning process can be solved by Recurrent Neural Networks (RNN) (i.e. PixelRNN, PixelCNN)\cite{oord2016pixel}. 
However, a single model of PixelRNN (or PixelCNN) maintains a fixed permutation $\pi$, and 
the permutation's order $\pi$ matters since it affects the structural constraints. 
Meanwhile, we need a different permutation $\pi^i$ for each condition $q_s(\mathbf{\hat{y}}_s^{\setminus i} | \hat{y}^i_s)$ according to the given conditional position on the segmentation.
Therefore, straightforwardly adopting RNN-based models as in prior works for  
all possible different permutations $\pi^i$ to compute $q_s(\mathbf{y}_s^{\setminus i} | y_s^i)$ requires a burdensome training process with numerous models (i.e. one model for each permutation) which is infeasible.

\begin{figure*}[!t]
    \centering
    \includegraphics[width=1.0\textwidth]{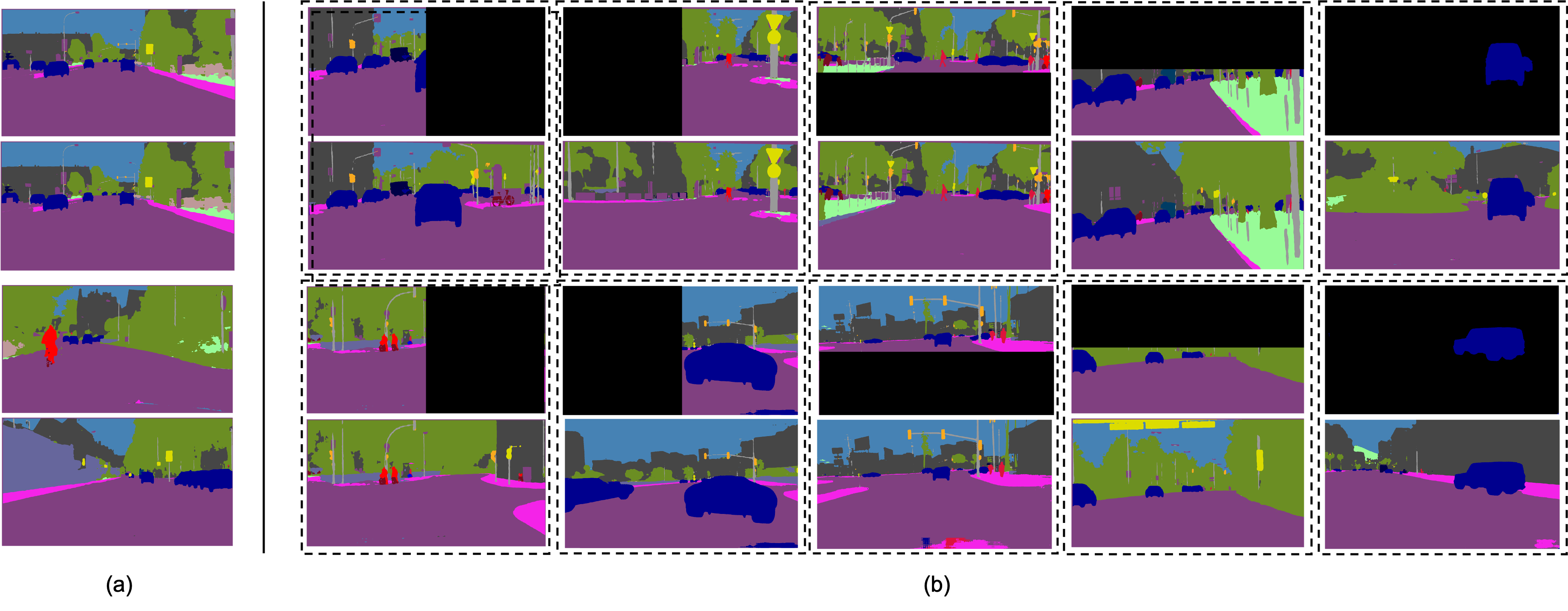}
    \caption{\textbf{Segmentation images sampled by the conditional structure network $G$.} (a) Fully Sampled Images. (b) Sampled Images Conditioned on a Given Mask.}
    \label{fig:gpt_sample}
\end{figure*}

Rather than learning numerous models for all possible orders $\pi^i$, we propose to represent $\mathbf{y}_s$ as sequence of pixels and mask out the unknown pixels. Then, the model is trained to maximize the conditional density function of these unknown pixels given the known pixels. As a results, by varying the input masks, the model can effectively obtain the capability of  learning the structure of the segmentation map $\mathbf{y}_s$ given known pixels. 
Formally, let $\mathbf{m} \in \mathcal{M}$ be a binary mask where value of \textit{one} indicates an unknown pixel and \textit{zero} is a known pixel, so the problem of learning the structural condition can be formulated as follows:
\begin{equation} \label{eqn:BERT_obj}
\small
\begin{split}
    \theta_G^* &= -\arg\min_{\theta_G} \mathbb{E}_{\mathbf{y} \sim \mathcal{Y}, \mathbf{m} \sim \mathcal{M}}\log p(\mathbf{y}_s \odot \mathbf{m} | \mathbf{y}_s \odot (1 - \mathbf{m}))
\end{split}
\end{equation}
where $\odot$ denotes the Hadamard product, $\mathbf{m}^i = 1$ means that the pixel $\mathbf{\hat{y}}_s^i$ is the masked pixel and vice versa.

Eqn. \eqref{eqn:BERT_obj} brings some interesting properties. \textit{Firstly}, if the $\mathbf{m}$ only contains one unmasked pixel, i.e., $\mathbf{m}^i = 0$ and $\forall j \in \{1..N\}, j \neq i: \mathbf{m}^j = 1$, the condition $p(\mathbf{y}_s \odot \mathbf{m} | \mathbf{y}_s \odot (1 - \mathbf{m}))$ is equivalent to $q_s(\mathbf{y}_s^{\setminus i} | y^i_s)$. %
Under this condition, 
the network $G$ %
can capture structural information of the segmentation conditioned on the given pixel.
\textit{Secondly}, 
if the binary mask only contains $\mathbf{1}$, i.e. $\mathbf{m}=\boldsymbol{1}$, the condition $p(\mathbf{y}_s \odot \mathbf{m} | \mathbf{y}_s \odot (1 - \mathbf{m}))$ is equivalent to the likelihood of 
$p(\mathbf{y}_s)$. 
Then, 
the network learns the global structure and the spatial relationship among pixels. 
\textit{Thirdly}, 
$\mathbf{m}$ is not only limited to one unmasked pixel but also able to contain more than one unmasked pixel,
it increases the flexibility of the model to learn the structures conditioned on the unmasked pixels.

To effectively learn the generative network $G$, we design it as a multi-head attention network, 
in which the spatial relationship and structure information can be learned by the attention mechanism at the training time. Particularly, considering each single pixel as a separate token,
we form the generator network $G$ with $L$ blocks of multihead attention network as follows:
\begin{equation}
\small
\begin{split}
    \mathbf{z}_0 &= \alpha(\mathbf{y}_s \odot (1-\mathbf{m})) + \beta(\boldsymbol{\pi}),  \quad \mathbf{a}_l = \mathbf{z}_l + \gamma(\tau(\mathbf{z})) \\ 
    \mathbf{z}_{l+1} &= \mathbf{a}_l + \eta(\tau(\mathbf{a}_l)), \quad\quad\quad\quad\quad\;\;\; \mathbf{\bar{y}}_s = \phi(\tau(\mathbf{z}_L)) \\
\end{split}
\end{equation}
where $\tau(\cdot)$ is the normalization norm \cite{layer_norm}, $\alpha(\cdot)$ is the token embedding network, $\beta(\boldsymbol{\pi})$ is the learned positional embedding ($\boldsymbol{\pi} = [1, 2, .., N]$),  
$\gamma(\cdot)$ represents the multi-head attention layer, $\eta(\cdot)$ is the residual-style multi-layer perception network, and $\phi(\cdot)$ is the projection that maps the normalized output of the multi-attention network
to the logits parameterizing the conditional distributions $p(\mathbf{y}_s \odot \mathbf{m} | \mathbf{y}_s \odot (1 - \mathbf{m})$. The subscript $l$ denotes the $l^{th}$ block. Fig. \ref{fig:framework}(b) illustrates the structural condition network. To ensure the correct structural conditions, 
we apply the proper binary mask into the $N \times N$ attention matrix.

\subsection{CoMaL properties}
\noindent
\textbf{Global Structural Learning.} Although these prior methods have achieved promising results for semantic segmentation, they usually adopted the pixel-independent assumption \cite{Chen_2019_ICCV, Araslanov:2021:DASAC}
and addressed long-tailed issues by predefining class-balanced weights. However, as shown in Eqn. \eqref{eqn:optimization_without_likelihood}, under the assumption of pixel independence, the condition $\frac{q'_s(\mathbf{y}^{\setminus i}_s| y^i_s)}{q_s(\mathbf{y}^{\setminus i}_s |y^i_s)}$ has been ignored. This results in lacking of mechanisms for maintain the structural learning during training process. Meanwhile, the conditional structural learning is explicitly formulated in the objective function of Eqn. \eqref{eqn:optimization_without_likelihood} and Section \ref{sec:ConditionalStructureLearning}, our CoMaL approach has shown its advantages in capturing the global structure of $\mathbf{y}_s$ and $\mathbf{y}_t$.

\noindent
\textbf{Class-balance Learning for both Source and Target Domains.} \textbf{Unlike} prior works where only labeled part is on their focus, CoMaL framework addresses the \textit{\textbf{long-tail}} issue in \textit{\textbf{both domains}} which is novel and more challenging. Experiments in Table \ref{tab:synthia2city} (a,b) have emphasized the advantages of CoMaL on improving the semantic segmentation accuracy on new unlabeled \mbox{domains}.

\noindent
\textbf{Flexibility Learning without the need of an Ideal Dataset.} CoMaL \textit{\textbf{does not require}} the presence of ideal data in its training procedure. As in Eqn. \eqref{eqn:CoMaL}, with any form of ideal distributions, the 
CoMaL %
loss is proven to be the upper bound of the loss obtained by ideal data. 
CoMaL also gives a flexibility in setting the desired class distribution $q’_s(\mathbf{y})$ even it is derived from non-uniform class distributions.

\noindent
\textbf{Relation to Markovian assumption.}
Rather than extracting neighborhood dependencies from local structures (i.e. pixels within a particular range) with Markovian assumption as in prior work, Eqn. \eqref{eqn:CoMaL}, $q_s(\mathbf{y}_s^{\setminus i} | y_s^i)$ emphasizes more on generalized structural constraints as it embeds dependencies on all remaining pixels of the segmentation map.
Particularly, after learning the conditional distribution $q_s(\mathbf{y}_s^{\setminus i} | y_s^i)$ via our proposed multihead attention network, during the training process, this term is considered as a metric to measure how good a predicted segmentation maintains the structural consistency (i.e., relative structure among objects) in comparison to the distributions of actual segmentation maps.

\section{Experiments}

This work was evaluated on two standard large-scale benchmarks including SYNTHIA $\to$ Cityscapes, GTA5 $\to$ Cityscapes. 
We firstly overview the datasets and network architectures in our experiments. 
Then, the ablation studies will be presented to analyze the performance of CoMaL approach. Finally, we present the quantitative and qualitative results of CoMAL method compared to prior methods on two benchmarks.

\subsection{Dataset Overview and Implementation Details}

\textbf{SYNTHIA (SYNHIA-RAND-CITYSCAPES)} \cite{Ros_2016_CVPR} is a synthetic segmentation dataset  generated from a virtual world to aid in semantic segmentation  of urban settings. 
It containing $9,400$ pixel-level labelled RGB images. There are 16 common classes that overlap with Cityscapes used in the experiments. SYNTHIA is licensed under Creative Commons Attribution-NonCommercial-ShareAlike 3.0. 

\noindent
\textbf{GTA5} \cite{Richter_2016_ECCV}, registered under the MIT License,  is a collection of $24,966$ synthetic, densely labelled images with the resolutions of $1914 \times 1052$ pixels. The dataset is collected from a game engine with 33 class categories, using the communication between the game engine and the graphics hardware. 
There are 19 categories 
compatible with the Cityscapes \cite{cordts2016cityscapes} used in our experiments. 

\noindent
\textbf{Cityscapes} \cite{cordts2016cityscapes} is a dataset of real-world urban images with $3,975$ high-quality, semantic, dense pixel annotations of 30 object classes.
The dataset was developed to improve and expand the number of high quality, annotated datasets of urban environments. 
We use $2,495$ images for training and $500$ images for testing during our experiments. 
The license of Cityscapes is made freely available to academic and nonacademic entities for non-commercial purposes.

\begin{table}[!t]
    \footnotesize
    \centering
    \caption{\textbf{The Effectiveness of Our Proposed BiMaL and CoMaL Losses}} %
    \begin{tabular}{c|l|c|c}
        Method & \multicolumn{1}{c|}{Setting} & 
        \begin{tabular}{@{}c@{}} SYNTHIA \\ $\to$ Cityscapes \end{tabular} &  \begin{tabular}{@{}c@{}} GTA5 \\ $\to$ Cityscapes \end{tabular} \\
        \hline
        \hline
        ReNet-101 & $-$ & 33.7 & 36.6 \\
        \hline
        \hline
        BiMaL & $\mathcal{L}_{llk}$  & 43.9 & 45.7 \\
        BiMaL & $\mathcal{L}_{llk} + \tau$  & 46.2 & 47.3 \\
        
        \hline 
        CoMaL & $\mathcal{L}_{cls}$  & 54.7 & 55.4  
        \\
        CoMaL & $\mathcal{L}_{cls}+\mathcal{L}_{CoMaL}$  & \textbf{57.8} & \textbf{59.3}
    \end{tabular}
    \label{tab:loss_abl}
\end{table}

\begin{figure}[!b]
    \centering
    \includegraphics[width=0.5\textwidth]{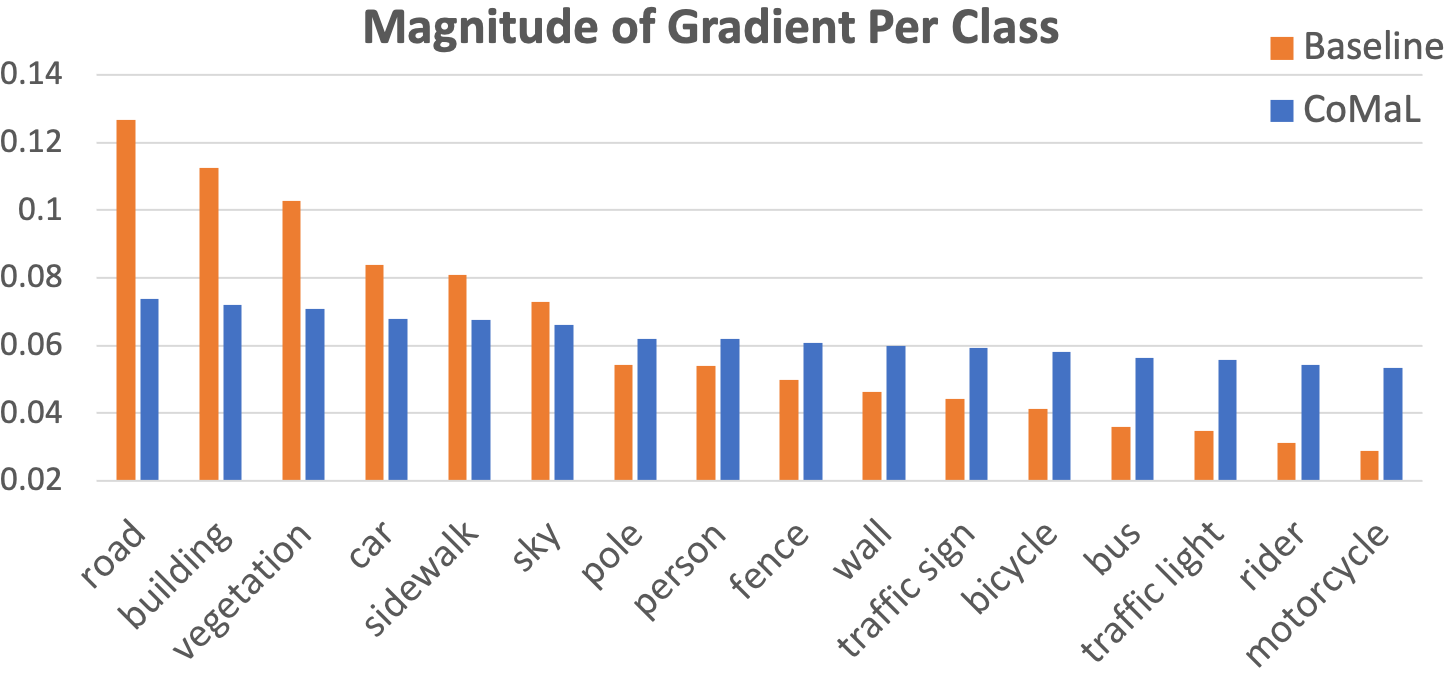}
    \caption{\textbf{Normalized Magnitude of Gradients per class.}}
    \label{fig:grad_per_class}
\end{figure}

\begin{table*}[!t]
        \centering
        \caption{\textbf{Semantic segmentation performance mIoU (\%) on validation set of Cityscapes with different approaches on the ResNet-101 backbone}}
            \setlength{\tabcolsep}{4pt}
            \begin{tabular}{c|c|c|c|c|c|c|c|c|c|c|c|c|c|c|c|c|c|c}
            \multicolumn{19}{c}{\textbf{SYNTHIA $\rightarrow$ Cityscapes (16 classes)}.  We also show the mIoU (\%) of the $13$ classes (mIoU*) excluding classes with *.} \\
				Models  & \rotatebox{90}{\textbf{road}} & \rotatebox{90}{\textbf{sidewalk}} & \rotatebox{90}{\textbf{building}} & \rotatebox{90}{\textbf{wall*}} & \rotatebox{90}{\textbf{fence*}} & \rotatebox{90}{\textbf{pole*}} & \rotatebox{90}{\textbf{light}} & \rotatebox{90}{\textbf{sign}} & \rotatebox{90}{\textbf{veg}} & \rotatebox{90}{\textbf{sky}} & \rotatebox{90}{\textbf{person}} & \rotatebox{90}{\textbf{rider}} & \rotatebox{90}{\textbf{car}} & \rotatebox{90}{\textbf{bus}} & \rotatebox{90}{\textbf{mbike}} & \rotatebox{90}{\textbf{bike}} & \rotatebox{90}{\textbf{mIoU}} & \rotatebox{90}{\textbf{mIoU*}}\\
				\hline \hline 
				ResNet-101 & \multirow{2}{*}{64.9} & \multirow{2}{*}{26.1} & \multirow{2}{*}{71.5} & \multirow{2}{*}{3.0} & \multirow{2}{*}{0.2} & \multirow{2}{*}{21.7} & \multirow{2}{*}{0.1} & \multirow{2}{*}{0.2} & \multirow{2}{*}{73.1} & \multirow{2}{*}{71.0} & \multirow{2}{*}{48.4} & \multirow{2}{*}{20.7} & \multirow{2}{*}{62.9} & \multirow{2}{*}{27.9} & \multirow{2}{*}{12.0} & \multirow{2}{*}{35.6} & \multirow{2}{*}{33.7} & \multirow{2}{*}{39.6} \\
				(without adaptation) & & & & & & & & & & & & & & & & & &\\ 
				\hline
				SPIGAN~\cite{lee2018spigan}&71.1&29.8&71.4&3.7&0.3&33.2&6.4&{15.6}&81.2&78.9&52.7&13.1&75.9&25.5&10.0&20.5&36.8&42.4\\
				AdaptPatch~\cite{tsai2019domain}&82.2&39.4&79.4&-&-&-&6.5&10.8&77.8&82.0&54.9&21.1&67.7&30.7&17.8&32.2&-&46.3\\
				CLAN~\cite{luo2018taking}&81.3&37.0&80.1&-&-&-&{16.1}&13.7&78.2&81.5&53.4&21.2&73.0&32.9&{22.6}&30.7&-&47.8\\
				AdvEnt~\cite{vu2019advent}&87.0&44.1&79.7&{9.6}&{0.6}&24.3&4.8&7.2&80.1&83.6&{56.4}&{23.7}&72.7&32.6&12.8&33.7&40.8&47.6\\
				IntraDA \cite{pan2020unsupervised} & 84.3 & 37.7 & 79.5 & 5.3 & 0.4 & 24.9 & 9.2 & 8.4 & 80.0 & 84.1 & {57.2} & 23.0 & 78.0 & 38.1 & 20.3 & 36.5 & 41.7 & 48.9 \\
				DADA\cite{vu2019dada} &{89.2}&{44.8}&{81.4}&6.8&0.3&26.2&8.6&11.1&{81.8}&{84.0}&54.7&19.3&{79.7}&{40.7}&14.0&{38.8}&{42.6}&{49.8} \\
				\textbf{BiMaL} & \textbf{92.8} & \textbf{51.5} & 81.5 & 10.2 & 1.0 & 30.4 & 17.6 & 15.9 & 82.4 & 84.6 & 55.9 & 22.3 & 85.7 & 44.5 & 24.6 & 38.8 & 46.2 & 53.7 \\
				\hline
				MaxSquare \cite{Chen_2019_ICCV} & 82.9 & 40.7 & 80.3 & 10.2 & 0.8 & 25.8 & 12.8 & 18.2 & 82.5 & 82.2 & 53.1 & 18.0 & 79.0 & 31.4 & 10.4 & 35.6 & 41.4 & 48.2 \\ 
				SAC \cite{Araslanov:2021:DASAC} & 89.3 & 47.3 & 85.6 & 26.6 & 1.3 & 43.1 & 45.6 & 32.0 & 87.1 & 89.1 & 63.7 & 25.3 & 87.0 & 35.6 & 30.3 & 52.8 & 52.6 & 59.3 \\
				\textbf{CoMaL} & 90.5 & 50.3 & \textbf{85.9} & \textbf{36.0} & \textbf{2.6} & \textbf{45.8} & \textbf{55.3} & \textbf{35.9} & \textbf{89.9} & \textbf{89.9} & \textbf{73.9} & \textbf{28.5} & \textbf{88.4} & \textbf{54.6} & \textbf{42.5} & \textbf{55.3} & \textbf{57.8} & \textbf{64.7} 
			\end{tabular}
			
			\label{tab:synthia2city} 
    \begin{tabular}{c|c|c|c|c|c|c|c|c|c|c|c|c|c|c|c|c|c|c|c|c}
    \multicolumn{21}{c}{\textbf{GTA5 $\to$ Cityscapes (19 classes).}} \\
        Models & \rotatebox{90}{\textbf{road}} & \rotatebox{90}{\textbf{sidewalk}} & \rotatebox{90}{\textbf{building}} & \rotatebox{90}{\textbf{wall}} & \rotatebox{90}{\textbf{fence}} & \rotatebox{90}{\textbf{pole}} & \rotatebox{90}{\textbf{light}} & \rotatebox{90}{\textbf{sign}} & \rotatebox{90}{\textbf{\textbf{veg}}} & \rotatebox{90}{\textbf{terrain}} & \rotatebox{90}{\textbf{sky}} & \rotatebox{90}{\textbf{\textbf{person}}} & \rotatebox{90}{\textbf{rider}} & \rotatebox{90}{\textbf{car}}& \rotatebox{90}{\textbf{truck}} & \rotatebox{90}{\textbf{bus}} & \rotatebox{90}{\textbf{train}} & \rotatebox{90}{\textbf{mbike}} & \rotatebox{90}{\textbf{bike}} & \rotatebox{90}{\textbf{mIoU}} \\
        \hline \hline
        ResNet-101  & \multirow{2}{*}{75.8} & \multirow{2}{*}{16.8} & \multirow{2}{*}{77.2} & \multirow{2}{*}{12.5} & \multirow{2}{*}{21.0} & \multirow{2}{*}{25.5} & \multirow{2}{*}{30.1} & \multirow{2}{*}{20.1} & \multirow{2}{*}{81.3} & \multirow{2}{*}{24.6} & \multirow{2}{*}{70.3} & \multirow{2}{*}{53.8} & \multirow{2}{*}{26.4} & \multirow{2}{*}{49.9} & \multirow{2}{*}{17.2} & \multirow{2}{*}{25.9} & \multirow{2}{*}{{6.5}} & \multirow{2}{*}{25.3} & \multirow{2}{*}{36.0} & \multirow{2}{*}{36.6} \\
        (without adaptation) & & & & & & & & & & & & & & & & & & & & \\ 
        \hline
        ROAD~\cite{chen2018road}                    & 76.3 & 36.1 & 69.6 & 28.6 & 22.4 & {28.6} & 29.3 & 14.8 & 82.3 & 35.3 & 72.9 & 54.4 & 17.8 & 78.9 & 27.7 & 30.3 & 4.0 & 24.9 & 12.6 & 39.4 \\
        AdaptSegNet~\cite{tsai2018learning}         & 86.5 & 36.0 & 79.9 & 23.4 & 23.3 & 23.9 & {35.2} & 14.8 & 83.4 & 33.3 & 75.6 & 58.5 & 27.6 & 73.7 & 32.5 & 35.4 & 3.9 & 30.1 & 28.1 & 42.4 \\
        MinEnt~\cite{vu2019advent}                  & 84.2 & 25.2 & 77.0 & 17.0 & 23.3 & 24.2 & 33.3 & {26.4} & 80.7 & 32.1 & 78.7 & 57.5 & {30.0} & 77.0 & {37.9} & 44.3 & 1.8 & 31.4 & {36.9} & 43.1 \\
        AdvEnt~\cite{vu2019advent}                  & {89.9} & {36.5} & {81.6} & {29.2} & {25.2} & {28.5} & 32.3 & 22.4 & 83.9 & 34.0 & 77.1 & 57.4 & 27.9 & {83.7} & 29.4 & 39.1 & 1.5 & 28.4 & 23.3 & 43.8 \\
        IntraDA \cite{pan2020unsupervised}                         & {90.6} & 36.1 & {82.6} & {29.5} & 21.3 & 27.6 & 31.4 & 23.1 & {85.2} & {39.3} & {80.2} & {59.3} & 29.4 & {86.4} & 33.6 & {53.9} & 0.0 & {32.7} & {37.6} & {46.3} \\
        \textbf{BiMaL} & \textbf{91.2} & {39.6} & {82.7} & {29.4} & {25.2} & {29.6} & {34.3} & 25.5 & {85.4} & {44.0} & {80.8} & {59.7} & {30.4 }& {86.6 }& {38.5} & {47.6} & 1.2 & {34.0} & 36.8 & {47.3} \\ 
        \hline
        MaxSquare \cite{Chen_2019_ICCV} & 89.4 & 43.0 & 82.1 & 30.5 & 21.3 & 30.3 & 34.7 & 24.0 & 85.3 & 39.4 & 78.2 & 63.0 & 22.9 & 84.6 & 36.4 & 43.0 & 5.5 & 34.7 & 33.5 & 46.4 \\
        SAC \cite{Araslanov:2021:DASAC} & 90.3 & 53.9 & 86.6 & 42.5 & 27.4 & 45.1 & 48.6 & 42.9 & 87.5 & 40.2 & 86.0 & 67.6 & 29.7 & 88.5 & 49.0 & 54.6 & \textbf{9.8} & 26.6 & 45.1 & 53.8 \\
        \textbf{CoMaL} & 90.6 & \textbf{58.7} & \textbf{87.9} & \textbf{44.1} & \textbf{44.4} & \textbf{47.3} & \textbf{54.1} & \textbf{52.9} & \textbf{88.7} & \textbf{47.1} & \textbf{87.4} & \textbf{70.6} & \textbf{39.7} & \textbf{88.8} & \textbf{55.3} & \textbf{58.7} & 8.7 & \textbf{47.7} & \textbf{54.4} & \textbf{59.3} 
    \end{tabular}
    \label{tab:gta52city}
\end{table*}

\noindent
\textbf{Implementation} We use the DeepLab-V2 \cite{chen2018deeplab} architecture with a ResNet-101 \cite{he2015deep} backbone for 
the segmentation network. The Atrous Spatial Pyramid Pooling sampling rate was set to $\{6,12,18,24\}$. 
The output of \textit{conv5} was used to predict the segmentation. 
The image size is set to $1280 \times 720$. %
For the experiments with the Transformer backbone, we adopt the MiT-B4 encoder \cite{xie2021segformer} as the backbone similar to \cite{daformer}.
We adopt the structure of \cite{chen2020generative} for our conditional structure network $G$. 
PyTorch \cite{paszke2019pytorch} was used to implement the framework. We used 4 NVIDIA Quadpro P8000 GPUs with 48GB of VRAM each.
The Stocahsict Gradient Descent optimizer \cite{Bottou10large-scalemachine} was used to train the framework with learning rate $2.5 \times 10^{-4}$, momentum $0.9$, weight decay $10^{-4}$, and batch size of 4 per GPU. %

\begin{table*}[!hb]
        \centering
        \caption{\textbf{Semantic segmentation performance mIoU (\%) on validation set of Cityscapes with different approaches on the Transformer backbone.}}
            \setlength{\tabcolsep}{4pt}
            \begin{tabular}{c|c|c|c|c|c|c|c|c|c|c|c|c|c|c|c|c|c|c}
				Models  & \rotatebox{90}{\textbf{road}} & \rotatebox{90}{\textbf{sidewalk}} & \rotatebox{90}{\textbf{building}} & \rotatebox{90}{\textbf{wall*}} & \rotatebox{90}{\textbf{fence*}} & \rotatebox{90}{\textbf{pole*}} & \rotatebox{90}{\textbf{light}} & \rotatebox{90}{\textbf{sign}} & \rotatebox{90}{\textbf{veg}} & \rotatebox{90}{\textbf{sky}} & \rotatebox{90}{\textbf{person}} & \rotatebox{90}{\textbf{rider}} & \rotatebox{90}{\textbf{car}} & \rotatebox{90}{\textbf{bus}} & \rotatebox{90}{\textbf{mbike}} & \rotatebox{90}{\textbf{bike}} & \rotatebox{90}{\textbf{mIoU}} & \rotatebox{90}{\textbf{mIoU*}}\\
    \multicolumn{19}{c}{\textbf{SYNTHIA $\rightarrow$ Cityscapes (16 classes)}.  We also show the mIoU (\%) of the $13$ classes (mIoU*) excluding classes with *.} \\
				\hline \hline 
				Swin-S \cite{transda} & \multirow{2}{*}{30.6} & \multirow{2}{*}{26.1} & \multirow{2}{*}{42.9} & \multirow{2}{*}{3.8} & \multirow{2}{*}{0.1} & \multirow{2}{*}{25.9} & \multirow{2}{*}{32.3} & \multirow{2}{*}{15.6} & \multirow{2}{*}{80.3} & \multirow{2}{*}{70.7} & \multirow{2}{*}{60.5} & \multirow{2}{*}{8.2} & \multirow{2}{*}{69.0} & \multirow{2}{*}{30.3} & \multirow{2}{*}{11.2} & \multirow{2}{*}{12.3} & \multirow{2}{*}{32.5} & \multirow{2}{*}{37.7} \\
				(without adaptation) & & & & & & & & & & & & & & & & & &\\ 
                Swin-B \cite{transda} & \multirow{2}{*}{57.3} & \multirow{2}{*}{33.8} & \multirow{2}{*}{56.0} & \multirow{2}{*}{6.3} & \multirow{2}{*}{0.2} & \multirow{2}{*}{33.8} & \multirow{2}{*}{35.5} & \multirow{2}{*}{18.9} & \multirow{2}{*}{79.9} & \multirow{2}{*}{74.8} & \multirow{2}{*}{63.1} & \multirow{2}{*}{10.9} & \multirow{2}{*}{78.3} & \multirow{2}{*}{39.0} & \multirow{2}{*}{20.8} & \multirow{2}{*}{19.4} & \multirow{2}{*}{39.2} & \multirow{2}{*}{45.2} \\
                (without adaptation) & & & & & & & & & & & & & & & & & &\\ 
                \hline
				TransDA-S \cite{transda} & 82.1 & 40.9 & 86.2 & 25.8 & 1.0 & 53.0 & 53.7 & 36.1 & 89.2 & 90.3 & 68.0 & 26.2 & 90.9 & 58.4 & 41.2 & 45.4 & 55.5 & 62.2 \\
				TransDA-B \cite{transda} & 90.4 & \textbf{54.8} & 86.4 & 31.1 & 1.7 & \textbf{53.8} & \textbf{61.1} & 37.1 & \textbf{90.3} & \textbf{93.0} & 71.2 & 25.3 & \textbf{92.3} & \textbf{66.0} & 44.4 & 49.8 & 59.3 & 66.3 \\
				DAFormer \cite{daformer} & 84.5 & 40.7 & 88.4 & \textbf{41.5} & 6.5 & 50.0 & 55.0 & \textbf{54.6} & 86.0 & 89.8 & 73.2 & 48.2 & 87.2 & 53.2 & 53.9 & 61.7 & 60.9 & 67.4 \\
				\hline
				\textbf{BiMaL} & 86.9 & 48.1 & 88.4 & 40.2 & 8.2 & 50.2 & 55.2 & 52.7 & 85.5 & 91.0 & 72.2 & 48.7 & 87.4 & 60.3 & 54.8 & 61.7 & 62.0 & 68.7 \\
				\textbf{CoMaL} & \textbf{90.5} & {50.3} & \textbf{88.5} & {40.4} & \textbf{8.2} &  {50.3} & {55.3} & 52.8 &  {89.9} &  {91.1} & \textbf{73.9} &  \textbf{49.3} &  {88.4} &  {61.1} &  \textbf{55.4} &  \textbf{62.8} &  \textbf{63.0} & \textbf{69.9} 
			\end{tabular}
			\label{tab:synthia2city_daformer} 
    \centering
    \begin{tabular}{c|c|c|c|c|c|c|c|c|c|c|c|c|c|c|c|c|c|c|c|c}
    \multicolumn{21}{c}{\textbf{GTA5 $\to$ Cityscapes (19 classes).}} \\
        Models & \rotatebox{90}{\textbf{road}} & \rotatebox{90}{\textbf{sidewalk}} & \rotatebox{90}{\textbf{building}} & \rotatebox{90}{\textbf{wall}} & \rotatebox{90}{\textbf{fence}} & \rotatebox{90}{\textbf{pole}} & \rotatebox{90}{\textbf{light}} & \rotatebox{90}{\textbf{sign}} & \rotatebox{90}{\textbf{\textbf{veg}}} & \rotatebox{90}{\textbf{terrain}} & \rotatebox{90}{\textbf{sky}} & \rotatebox{90}{\textbf{\textbf{person}}} & \rotatebox{90}{\textbf{rider}} & \rotatebox{90}{\textbf{car}}& \rotatebox{90}{\textbf{truck}} & \rotatebox{90}{\textbf{bus}} & \rotatebox{90}{\textbf{train}} & \rotatebox{90}{\textbf{mbike}} & \rotatebox{90}{\textbf{bike}} & \rotatebox{90}{\textbf{mIoU}} \\
        \hline \hline
        Swin-S \cite{transda} & \multirow{2}{*}{55.9} & \multirow{2}{*}{21.8} & \multirow{2}{*}{63.1} & \multirow{2}{*}{14.0} & \multirow{2}{*}{22.0} & \multirow{2}{*}{27.2} & \multirow{2}{*}{46.8} & \multirow{2}{*}{17.4} & \multirow{2}{*}{83.3} & \multirow{2}{*}{32.8} & \multirow{2}{*}{86.1} & \multirow{2}{*}{62.2} & \multirow{2}{*}{28.7} & \multirow{2}{*}{43.8} & \multirow{2}{*}{32.2} & \multirow{2}{*}{36.9} & \multirow{2}{*}{1.1} & \multirow{2}{*}{34.8} & \multirow{2}{*}{35.5} & \multirow{2}{*}{39.2} \\
        (without adaptation) & & & & & & & & & & & & & & & & & & & & \\ 
        Swin-B \cite{transda} & \multirow{2}{*}{63.3} & \multirow{2}{*}{28.6} & \multirow{2}{*}{68.3} & \multirow{2}{*}{16.8} & \multirow{2}{*}{23.4} & \multirow{2}{*}{37.8} & \multirow{2}{*}{51.0} & \multirow{2}{*}{34.3} & \multirow{2}{*}{83.8} & \multirow{2}{*}{42.1} & \multirow{2}{*}{85.7} & \multirow{2}{*}{68.5} & \multirow{2}{*}{25.4} & \multirow{2}{*}{83.5} & \multirow{2}{*}{36.3} & \multirow{2}{*}{17.7} & \multirow{2}{*}{2.9} & \multirow{2}{*}{36.1} & \multirow{2}{*}{42.3} & \multirow{2}{*}{44.6} \\
        (without adaptation) & & & & & & & & & & & & & & & & & & & & \\ 
        \hline
        TransDA-S \cite{transda} & 92.9 & 59.1 & 88.2 & 42.5 & 32.0 & 47.6 & 57.6 & 39.2 & 89.6 & 42.0 & 94.1 & 74.3 & 45.3 & 91.4 & 54.0 & 58.0 & 44.4 & 48.3 & 51.4 & 60.6  \\
		TransDA-B \cite{transda} & 94.7 & 64.2 & 89.2 & 48.1 & 45.8 & 50.1 & \textbf{60.2} & 40.8 & \textbf{90.4} & \textbf{50.2} & \textbf{93.7} & \textbf{76.7} & \textbf{47.6} & 92.5 & 56.8 & 60.1 & 47.6 & 49.6 & 55.4 & 63.9 \\
        DAFormer \cite{daformer} & 95.7 & 70.2 & 89.4 & 53.5 & 48.1 & 49.6 & 55.8 & 59.4 & 89.9 & 47.9 & 92.5 & {72.2} & 44.7 & 92.3 & 74.5 & 78.2 & 65.1 & 55.9 & 61.8 & 68.3 \\
        \hline
        \textbf{BiMaL} & 96.1 & 72.0 & 89.3 & 52.4 & 48.1 & 48.7 & 56.2 & 61.0 & 89.5 & 46.4 & 92.3 & 72.0 & 46.1 & 91.9 & 67.4 & 80.9 & 67.7 & 54.4 & 61.7 & 68.1\\
        \textbf{CoMaL} & \textbf{96.7} & \textbf{74.8} & \textbf{89.5} & \textbf{55.2} & \textbf{48.3} & \textbf{50.6} & {56.3} & \textbf{63.6} & {90.0} & {49.0} & {92.6} & 72.0 & {46.3} & \textbf{92.6} & \textbf{80.1} & \textbf{81.0} & \textbf{70.1} & \textbf{57.7} & \textbf{63.4} & \textbf{70.0} \\
       
    \end{tabular}
    \label{tab:gta52city_daformer}
\end{table*}

\begin{figure*}[t]
    \centering
    \includegraphics[width=1.0\textwidth]{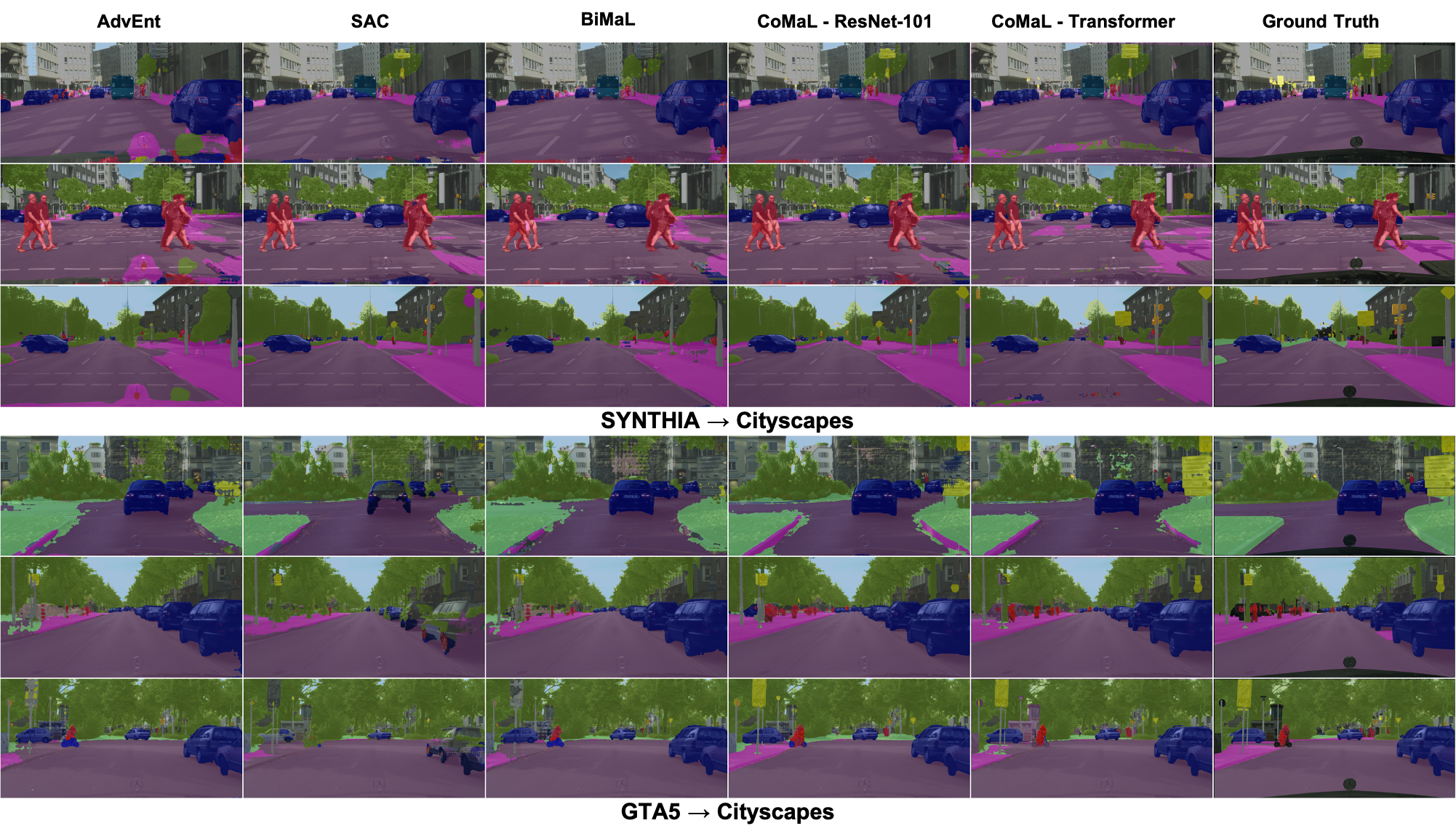}
    \caption{\textbf{Qualitative Results.} %
    We compare our results, i.e., BiMaL, CoMaL with the ResNet-101 backbone, and CoMal with the Transformer backbone, with 
    AdvEnt \cite{vu2019advent}, SAC \cite{Araslanov:2021:DASAC},  
    (a) \textbf{SYNTHIA $\to$ Cityscapes} and (b) \textbf{GTA5 $\to$ Cityscapes} (Best view in color).
    }
    \label{fig:qualtitave_rec_syn2city}
\end{figure*}

\begin{figure*}[t]
    \centering
    \includegraphics[width=1.0\textwidth]{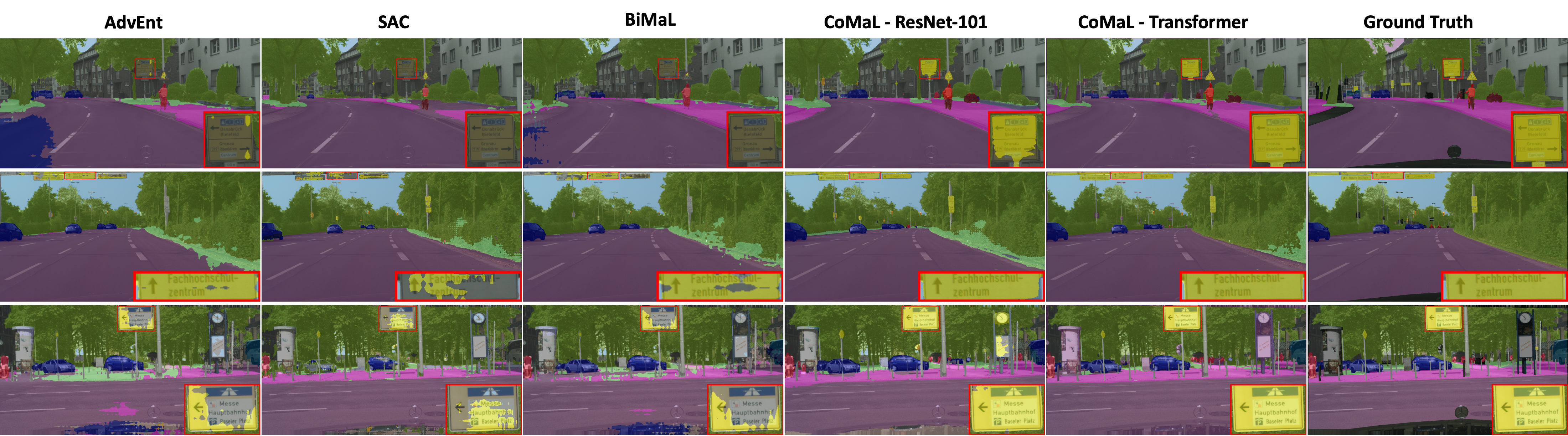}
    \caption{\textbf{Qualitative Results of Long-tail Classes.} We highlight the long-tail class (i.e., traffic sign) and compare the results of AdvEnt \cite{vu2019advent}, SAC \cite{Araslanov:2021:DASAC}, BiMaL, CoMaL with the ResNet-101 backbone, and CoMaL with the Transformer backbone.} 
    \label{fig:highlight_long_tail_vis}
\end{figure*}
  
\subsection{Ablation Study}

\noindent
\textbf{Structural Learning with Conditional Structure Network}
This experiment illustrates the capability of the structural condition modeling of the generative network $G$ trained on GTA5.  
As in Fig. \ref{fig:gpt_sample}(a), our structural condition network can capture the structure information of the segmentation. In particular, at the detail level of the segmentation, the structural borders between objects are well modeled. Moreover, as in Fig. \ref{fig:gpt_sample}(b), our conditional network $G$ is able to model the segmentation conditioned on the given objects.
This experiment has confirmed that our approach can sufficiently model the segmentation even with complex and diverse structures as scene segmentation.

\noindent
\textbf{Effectiveness of Losses} 
Table \ref{tab:loss_abl} reports the mIoU performance of our proposed BiMaL and CoMaL losses compared to the baseline of ResNet-101.
In BiMaL experiments, we consider three cases: (1) without adaptation (train with source only), (2) BiMaL without regularization term ($\mathcal{L}_{llk}(\mathbf{y})$ only), and (3) BiMaL with regularization term ($\mathcal{L}_{llk}(\mathbf{y}) + \tau(\mathbf{y})$). 
Overall, the proposed BiMaL improve the performance of the method. In particular, the mIoU accuracy of the baseline (without adaptation) is $33.7\%$ and $36.6\%$. 
In comparison, BiMaL without regularization and BiMaL with regularization achieve the mIoU accuracy of $43.5\%$ and $46.2\%$ on SYNTHIA $\to$ Cityscapes, and $46.2\%$ and $47.3\%$ on GTA5 $\to$ Cityscapes, respectively. 
In CoMaL experiments, we deploy only the conditional class-weighted loss ($\mathcal{L}_{cls}$), the performance of CoMaL are improved up to $54.7\%$ and $55.4\%$ on the benchmarks of SYNTHIA $\to$ Cityscapes and GTA5 $\to$ Cityscapes, respectively. 
Moreover, when we utilize both class-weighted loss and conditional maximum likelihood loss ($\mathcal{L}_{CoMaL}$), the performance of our proposals improves by a large margin and achieves the mIoU up to $57.8\%$ on SYNTHIA $\to$ Cityscapes and $59.3\%$ on GTA5 $\to$ Cityscapes.

\noindent
\textbf{Effectiveness of Gradients}
We visualize the effect of gradients updated to each class by 
taking a subset of the validation set of Cityscapes to compute and average the gradients of predictions with respect to each class. 
Fig. \ref{fig:grad_per_class} illustrates the normalized magnitude of gradients with respect to each class. 
The gradients of each class produced by our method have fairly updated with the predictions. . 
Meanwhile, without CoMaL, the gradients updated to the predictions of head class largely dominate the tailed classes.

\subsection{Quantitative and Qualitative Results}

In this experiment, we compare our CoMaL %
with prior SOTA methods on 
two benchmarks: SYNTHIA $\to$ Cityscapes and GTA5 $\to$ Cityscapes. 
As other standard benchmarks \cite{vu2019advent, Araslanov:2021:DASAC}, the performance of segmentation is compared by 
the mean Intersection over Union (mIoU) metric.

\textbf{SYNTHIA $\to$ Cityscapes:} 
Table \ref{tab:synthia2city} presents the our SOTA performance of the proposed BiMaL and CoMaL approach compared to prior methods on the ReNet-101 backbone over 16 common classes on the Cityscape validation set. 
In the BiMaL experimental setting, we have compared our BiMaL results with prior approaches in which the long-tail issue has been not aware.
Our proposed BiMaL achieves better accuracy than the prior methods, i.e. $46.2\%$ higher than DADA \cite{vu2019dada} by $3.6\%$. Considering per-class results, our method significantly improves the results on classes of \textit{`sidewalk'} ($51.5\%$), \textit{`car'} ($85.7\%$), and \textit{`bus'} ($44.5\%$). 
Meanwhile, in the experimental setting of CoMaL in which the long-tail issue has been taken in to account, 
our CoMaL approach achieved higher than the previous SOTA method (SAC) \cite{Araslanov:2021:DASAC} by $5.2\%$.
Considering the per class result, %
in comparison with SAC \cite{Araslanov:2021:DASAC}, our method can significantly improve the results on the tailed classes, e.g. \textit{``bus''} ($+19.0\%$), \textit{``motorbike''} ($+12.2\%$), \textit{``person''} ($+10.2\%$), \textit{``traffic light''} ($+9.7\%$), \textit{``wall''} ($+9.4\%$), etc.
In addition, our approach maintains and improves the performance of the head class compared to SAC \cite{Araslanov:2021:DASAC}, 
i.e  \textit{``road''} ($90.5\%$), \textit{``building''} ($85.9\%$), \textit{``vegetation''} ($89.9\%$), \textit{``car''} ($88.4\%$), \textit{``sidewalk''} ($50.3\%$), \textit{``sky''} ($89.9\%$).
We also report the results on a 13-class subset where our proposed method also achieves the State-of-the-Art performance.
Additionally, we have reported our results trained on Transformer backbone as reported in Table \ref{tab:synthia2city_daformer}. In this experiment, we also achieve the state-of-the-art performance compared with prior approaches trained on Transformer backbone.

\textbf{GTA5 $\to$ Cityscapes:}
Table \ref{tab:gta52city} illustrates the mIoU accuracy over 19 classes of Cityscapes. %
Overall, our CoMaL approach achieves the mIoU accuracy of $59.3\%$ that is the SOTA performance compared to the prior methods on the same ResNet-101 backbone.
In the BiMaL experimental setting, our BiMaL approach achieved mIoU of $47.3\%$ that is SOTA performance compared to the previous methods. 
Analysing per-class results, our method gains the improvement on most classes, e.g., the results on classes of \textit{`terrain'} (+$10.0\%$), \textit{`truck'} ($+9.1\%$), \textit{`bus'} ($+8.0\%$), \textit{`motorbike'} ($+5.6\%$) show significant improvements compared to AdvEnt. 
Meanwhile, in the CoMaL experimental setting, further analysis suggests that our method achieves comparable results and improves the performance of segmentation on the tail classes. 
In particular, the results on the tail classes of  \textit{``terrain''}, \textit{``fence''}, \textit{``traffic sign''}, \textit{``bicycle''}, \textit{``truck''}, \textit{``rider''}, and \textit{``motorcycle''} have been improved by $6.9\%$, $17.0\%$, $10.0\%$, $9.3\%$, $6.3\%$, $10.0\%$, and $21.1\%$, respectively.
Furthermore, the performance of the head classes are maintained and improved, i.e. the mIoU results of \textit{``road''}, \textit{``building''}, \textit{``vegetation''}, \textit{``car''},  and \textit{``sky''} are $90.6\%$, $87.9\%$, $88.7\%$, $88.8\%$, and $87.4\%$, respectively. 
Also, we have achieved the state-of-the-art performance and compared our results with prior approaches trained on the Transformer backbone as reported in Table \ref{tab:gta52city_daformer}.

\textbf{Qualitative Results:} As shown in Fig. \ref{fig:qualtitave_rec_syn2city}, 
our approach produces better qualitative results compared to prior SOTA approaches,
, i.e. AdvEnt \cite{vu2019advent}, BiMaL , DA-SAC \cite{Araslanov:2021:DASAC}, BiMaL, CoMaL with the ResNet-101 backbone, and CoMaL with the Transformer backbone. 
The most distinguishable improvements come from the tail classes that occupy smaller portions of a given image. 
In particular, the segmentation of instances occupying tail classes such as signs, people, and poles are more clearly defined. 
The model is able to accurately identify the border regions of these classes. 
Figure \ref{fig:highlight_long_tail_vis} highlights our segmentation results on long-tail classes compared to other approaches.
The continuity of each instance more clearly matches the ground truth labels, as the model does not leave out as many parts of the smaller objects. 
It is able to cohesively segment the tail classes, reducing the portion of each tail-class that is incorrectly labeled, while also recognizing where it begins and ends. 
Moreover, the head classes are qualitatively maintained and enhanced along with the tail classes. 
Although there is little noise in the head classes, the boundaries are still precise and match the ground truth. 

\section{Conclusions}
This paper has presented a novel metric for cross-domain long-tail adaptation in semantic segmentation. 
A new structural condition network has been introduced to learn the spatial relationship and structure information of the segmentation. Further, conditional maximum likelihood losses have been derived to tackle the issues of long-tail distributions that occur in both domains. The intensive experiments on two  benchmarks, i.e., SYNTHIA $\to$ Cityscapes, GTA $\to$ Cityscapes, have shown the performance of our CoMaL approach. 
The method achieves the SOTA performance in both benchmarks and improves the performance of the segmentation on tailed class, while also maintaining and improving the mIoU results on head classes.

\bibliographystyle{IEEEtran}
\bibliography{references}

\begin{IEEEbiography}[{\includegraphics[width=1in,height=1.25in,clip,keepaspectratio]{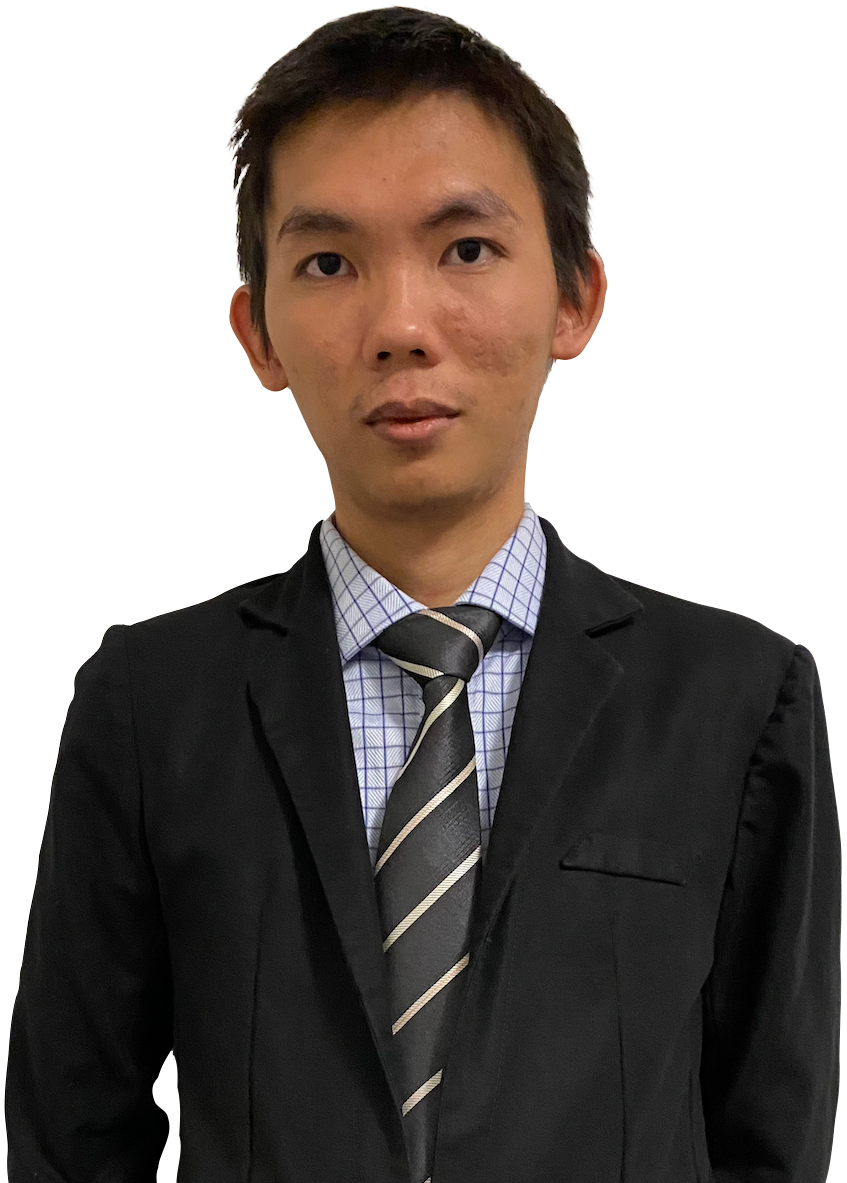}}]{Thanh-Dat Truong}
is currently a Ph.D. Candidate at the Department of Computer Science and Computer Engineering of the University of Arkansas. He received his B.Sc. degree in Computer Science from Honors Program, University of Science, VNU in 2019. He was a research intern at Coordinated Lab Science at the University of Illinois at Urbana-Champaign in 2018. When Thanh-Dat Truong was an undergraduate student, he worked as a research assistant at Artificial Intelligence Lab at the University of Science, VNU. Thanh-Dat Truong's research interests widely include Face Recognition, Action Recognition, Domain Adaptation, Deep Generative Model, Adversarial Learning. His papers appear at top tier conferences such as Computer Vision and Pattern Recognition, International Conference on Computer Vision , International Conference on Pattern Recognition, Canadian Conference on Computer and Robot Vision. 
He is also a reviewer of top-tier journals and conferences including IEEE Transaction on Pattern Analysis and Machine Intelligence, 
IEEE Transaction on Image Processing,
Journal of Computers Environment and Urban Systems, 
IEEE Access, 
Computer Vision and Pattern Recognition,
European Conference on Computer Vision,
Asian Conference on Computer Vision,
Winter Conference on Applications of Computer Vision, 
International Conference on Pattern Recognition.
\end{IEEEbiography}


\begin{IEEEbiography}[{\includegraphics[width=1in,height=1.25in,clip,keepaspectratio]{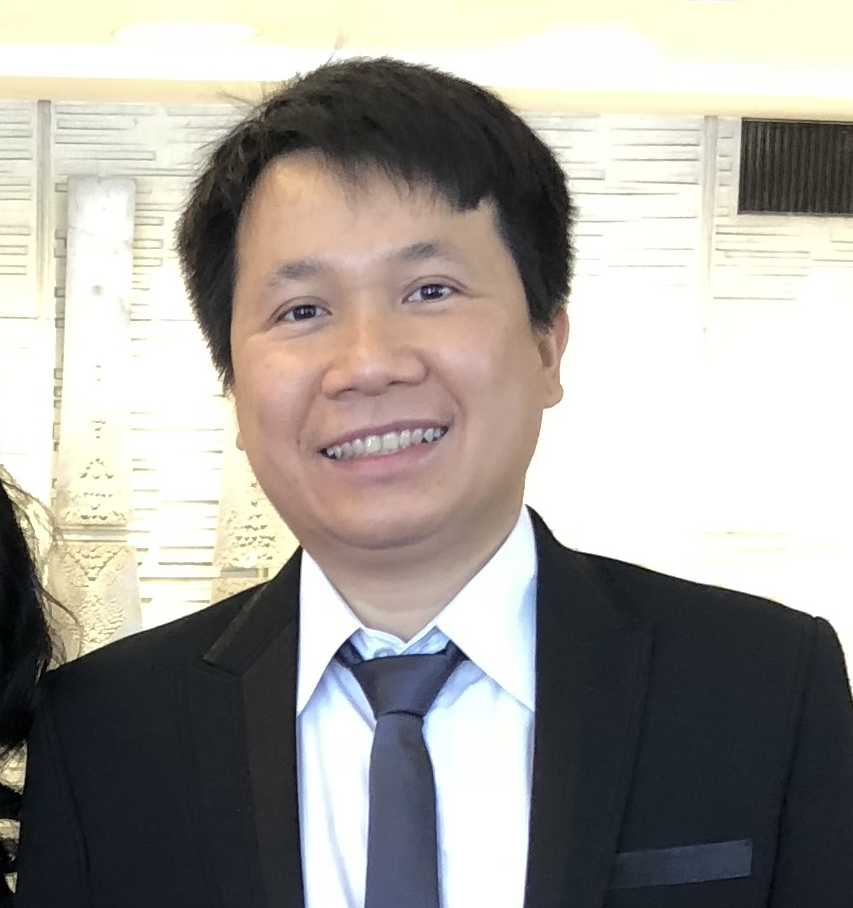}}]{Chi Nhan Duong}
is currently a Senior Technical Staff and having research collaborations with both Computer Vision and Image Understanding (CVIU) Lab, University of Arkansas, USA and Concordia University, Montreal, Canada. He had been a Research Associate in Cylab Biometrics Center at Carnegie Mellon University (CMU), USA since September 2016. He received his Ph.D. degree in Computer Science with the Department of Computer Science and Software Engineering, Concordia University, Montreal, Canada. He was an Intern with National Institute of Informatics, Tokyo Japan in 2012. He received his B.S. and M.Sc. degrees in Computer Science from the Department of Computer Science, Faculty of Information Technology, University of Science, Ho Chi Minh City, Vietnam, in 2008 and 2012, respectively. His research interests include Deep Generative Models, Face Recognition in surveillance environments, Face Aging in images and videos, Biometrics, and Digital Image Processing, and Digital Image Processing (denoising, inpainting and super-resolution). He is currently a reviewer of several top-tier journals including IEEE Transaction on Pattern Analysis and Machine Intelligence (TPAMI), IEEE Transaction on Image Processing (TIP), Journal of Signal Processing, Journal of Pattern Recognition, Journal of Pattern Recognition Letters. He is also recognized as an outstanding reviewer of several top-tier conferences such as The IEEE Computer Vision and Pattern Recognition (CVPR), International Conference on Computer Vision (ICCV), European Conference On Computer Vision (ECCV), Conference on Neural Information Processing Systems (NeurIPS), International Conference on Learning Representations (ICLR) and the AAAI Conference on Artificial Intelligence.
He is also a Program Committee Member of Precognition: Seeing through the Future, CVPR.
\end{IEEEbiography}

\begin{IEEEbiography}[{\includegraphics[width=1in,height=1.25in,clip,keepaspectratio]{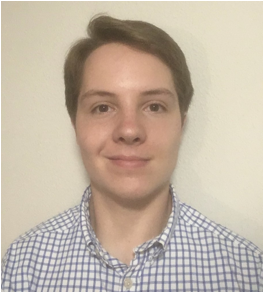}}]{Pierce Helton}
is an undergraduate student in the Department of Computer Science and Computer Engineering at the University of Arkansas. He is expected to graduate in Fall of 2022. He is currently working in the Computer Vision and Image Understanding lab at his university. His research interests are Machine Learning, Deep Learning, and Computer Vision. Pierce has plans to work in the industry and potentially pursue a M.S. in Computer Science after earning his B.S.
\end{IEEEbiography}

\begin{IEEEbiography}[{\includegraphics[width=1in,height=1.25in,clip,keepaspectratio]{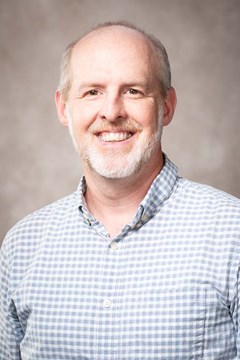}}]{Ashley Dowling} is currently a Professor in the Department of Entomology and Plant Pathology at the University of Arkansas. He is serving as Editor-in-Chief of the International Journal of Acarology. His research interests focus on biodiversity, evolutionary biology, and ecology of insects and other arthropods. He has coauthored 80+ papers in journals on these topics and trained more than 20 graduate students. 
\end{IEEEbiography}

\begin{IEEEbiography}[{\includegraphics[width=1in,height=1.25in,clip,keepaspectratio]{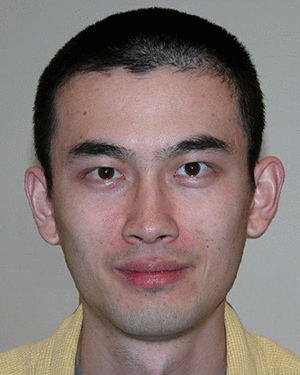}}]{Xin Li} (Fellow, IEEE) received the B.S. (Hons.) degree in electronic engineering and information science from the University of Science and Technology of China, Hefei, China, in 1996, and the Ph.D. degree in electrical engineering from Princeton University, Princeton, NJ, USA, in 2000. From 2000 to 2002, he was a Technical Staff Member with the Sharp Laboratories of America, Camas, WA, USA. Since 2003, he has been a Faculty Member with the Lane Department of Computer Science and Electrical Engineering. His research interests include image/video coding and processing. He was the recipient of the Best Student Paper Award at the Conference of Visual Communications and Image Processing in 2001, Best Student Paper Award at the IEEE Asilomar Conference on Signals, Systems and Computers in 2006, and Best Paper Award at the Conference of Visual Communications and Image Processing in 2010. He is currently a Member of the Image, Video, and Multidimensional Signal Processing Technical Committee and an Associate Editor for the IEEE Transactions on Circuits and Systems for Video Technology.
\end{IEEEbiography}

\begin{IEEEbiography}[{\includegraphics[width=1in,height=1.25in,clip,keepaspectratio]{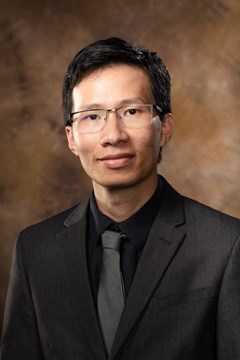}}]{Khoa Luu}
is currently an Assistant Professor and the Director of Computer Vision and Image Understanding (CVIU) Lab
in Department of Computer Science \& Computer Engineering at University of
Arkansas. He is serving as an Associate Editor of IEEE Access journal. He was the Research Project Director in Cylab Biometrics Center at Carnegie Mellon University (CMU), USA. He has received four patents and two best paper awards, and coauthored 120+ papers in conferences and journals. He was a vice chair of Montreal Chapter IEEE SMCS in Canada from September
2009 to March 2011. His research expertise includes Biometrics, Face Recognition, Tracking, Human Behavior Understanding, Scene Understanding, Domain Adaptation, Deep Generative Modeling, Image and Video Processing, Deep Learning, Compressed Sensing and Quantum Machine Learning. 
He is a co-organizer and a chair of CVPR Precognition Workshop in 2019, 2020, 2021, 2022; MICCAI Workshop in 2019, 2020 and ICCV Workshop in 2021. He is a PC member of AAAI, ICPRAI in 2020, 2022. He has been an active reviewer for several AI conferences and journals, such as CVPR, ICCV, ECCV, NeurIPS, ICLR, IEEE-TPAMI, IEEE-TIP, IEEE Access, Journal of Pattern Recognition, Journal of Image and Vision Computing, Journal of Signal Processing, and Journal of Intelligence Review.
\end{IEEEbiography}




\end{document}